\documentclass[11pt]{article}

\usepackage[final]{acl}

\usepackage{times}
\usepackage{latexsym}

\usepackage[T1]{fontenc}

\usepackage[utf8]{inputenc}

\usepackage{microtype}

\usepackage{inconsolata}

\usepackage{graphicx}
\usepackage{algorithm}
\usepackage{graphicx}
\usepackage{amsmath} 
\usepackage{amssymb} 
\usepackage{booktabs}
\usepackage{multirow}
\usepackage{float} 
\usepackage{amsmath}
\usepackage{newfloat}
\usepackage{listings}
\usepackage{algpseudocode}
\usepackage{xcolor}
\usepackage{amssymb}
\usepackage[normalem]{ulem}
\usepackage{hyperref}
\usepackage{fontawesome5}
\useunder{\uline}{\ul}{}
\title{Data Selection for Multi-turn Dialogue Instruction Tuning}

\author{
Bo Li\textsuperscript{\rm 1,2,3},
Shikun Zhang\textsuperscript{\rm 1},
Wei Ye\textsuperscript{\rm 1}\thanks{Corresponding author}
\\
\textsuperscript{\rm 1}National Engineering Research Center for Software Engineering, Peking University \\
\textsuperscript{\rm 2}School of Computer Science, Peking University \\
\textsuperscript{\rm 3}PKU-CMCC(Hubei) Joint Research Lab for LLM Industrial Applications\\
\texttt{deepblue.lb@gmail.com, wye@pku.edu.cn}\\
\faGithub\ \href{https://github.com/WisdomShell/MDS}{WisdomShell/MDS}
\faGlobe\ \href{https://wisdomshell.github.io/MDS/}{MDS Project}
}

\begin{document}
\maketitle
\begin{abstract}
Instruction-tuned language models increasingly rely on large multi-turn dialogue corpora, but these datasets are often noisy and structurally inconsistent, with topic drift, repetitive chitchat, and mismatched answer formats across turns.
We address this from a data selection perspective and propose \textbf{MDS} (Multi-turn Dialogue Selection), a dialogue-level framework that scores whole conversations rather than isolated turns. MDS combines a global coverage stage that performs bin-wise selection in the user-query trajectory space to retain representative yet non-redundant dialogues, with a local structural stage that evaluates within-dialogue reliability through entity-grounded topic grounding and information progress, together with query-answer form consistency for functional alignment. MDS outperforms strong single-turn selectors, dialogue-level LLM scorers, and heuristic baselines on three multi-turn benchmarks and an in-domain Banking test set, achieving the best overall rank across reference-free and reference-based metrics, and is more robust on long conversations under the same training budget. Code and resources are included in the supplementary materials.

\end{abstract}

\section{Introduction}
Supervised fine-tuning on instruction-style data is now a central step in turning base language models into aligned assistants, from RLHF to recent instruction-tuned open-source models~\cite{Ouyang2022TrainingLM,Wang2022SuperNaturalInstructionsGV,taori2023stanford,kopf2023openassistant,dubey2024llama,yang2025qwen3,tian-etal-2025-compkbqa,tian-etal-2025-grv,zhao2026generatingeffectivecottraces}. Yet a series of studies have shown that simply increasing dataset size is not sufficient and can even hurt downstream behavior when the supervision is noisy, redundant, or off-distribution~\cite{zhou2023lima,wang2023openchat,li2024quantity,li2026instructiondataselectionanswer}. Work on small, high-quality alignment sets consistently shows that data composition matters for shaping model behavior~\cite{qi2023fine,dong2024abilities,shen2024rethinking}, motivating instruction-data selection and reweighting methods that aim to identify the most beneficial supervision signals for downstream capabilities.

\begin{figure}[t]
    \centering
    \includegraphics[width=0.89\linewidth]{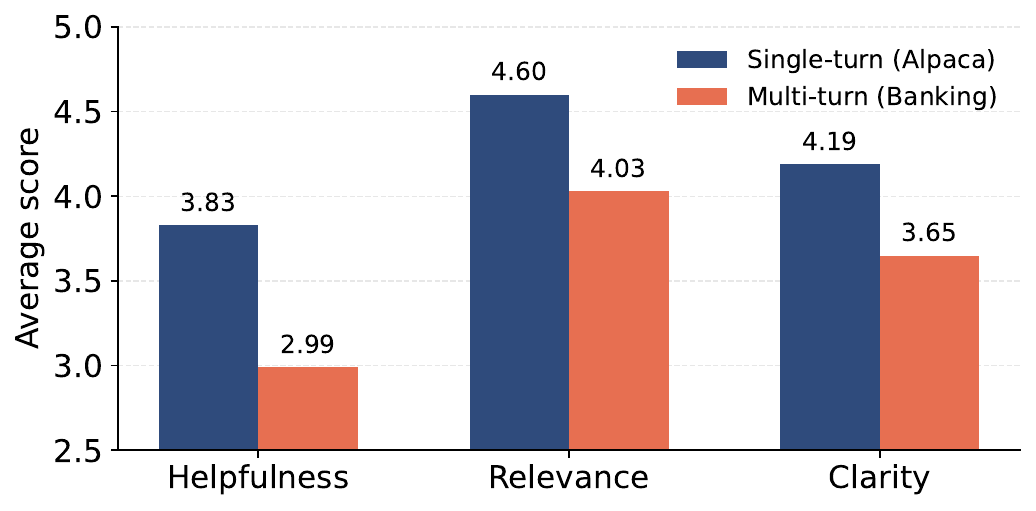}
    \caption{Comparison of turn-level quality between a single-turn instruction dataset (Alpaca) and a multi-turn dialogue dataset (Banking). We randomly sample 1,000 examples from each dataset and score the assistant responses for helpfulness, relevance and clarity on a 1–5 scale using GPT-4o as the evaluator. }
    \label{pic.data}
\end{figure}

Most existing work on data selection still focuses on single-turn instruction–response pairs, where examples are easy to synthesize and each instance can be scored in isolation. Recent methods select or reweight instructions based on self-guided signals, LLM quality scores, or simple heuristics such as response length, and have shown clear gains for instruction tuning~\cite{wang2023self,li2024superfiltering,liu2024selectit,xia2024less,xia2024rethinking,mekala2024smaller,he2025fine}. In contrast, multi-turn dialogue data are usually collected from human–assistant interaction logs or large-scale synthetic generators~\cite{ding2023enhancing,wang2023target,xu2023baize,zheng2023lmsys}, and we empirically find that their quality is often lower and more variable across turns. A simple turn-level comparison already makes this gap visible: Figure~\ref{pic.data} compares a standard single-turn instruction dataset (Alpaca~\cite{Peng2023InstructionTW}) with a multi-turn dialogue corpus (Banking) and shows that the latter consistently receives lower scores in helpfulness, relevance, and clarity on a 1–5 scale. Beyond lower average scores, manual inspection reveals that the multi-turn corpus suffers from characteristic dialogue-level failures: later turns often drift away from the user's original intent, many conversations end with long chitchat tails, and some responses ignore the requested format (e.g., open-ended advice instead of concrete steps). These issues are hard to detect from isolated turns but accumulate over trajectories, degrading the value of multi-turn supervision.
While several works have begun to construct large multi-turn corpora and analyze consistency~\cite{liu2023g,lin2023llm,chen2025consistentchat}, their processing pipelines still rely mainly on rule-based or coarse filtering. These gaps motivate us to develop a dialogue-level data selection method that explicitly targets multi-turn structure and conversational quality, rather than treating each turn as an independent single-turn instruction.

In this paper, we introduce \textbf{MDS} (\textbf{M}ulti-turn \textbf{D}ialogue \textbf{S}election), combining a \textit{global semantic coverage} stage with a \textit{local structural} stage.
In the global stage, MDS embeds each dialogue into a \emph{user-query trajectory} representation, which captures the evolving intent while being robust to assistant-side chitchat.
We then partition the trajectory space into semantic bins and perform \emph{bin-wise semantic coverage} selection within each bin using an efficient greedy coverage--redundancy criterion, yielding a representative yet non-redundant subset under a strict budget.
This global mechanism explicitly prevents a few high-frequency interaction patterns from dominating the selection and improves long-tail intent coverage.
In the local stage, MDS assesses \emph{within-dialogue structural reliability} by measuring entity-grounded \emph{topic grounding} and \emph{information progress} across turns, together with query-answer form consistency that enforces functional alignment between query types and response formats.
By prioritizing dialogues that are both well-covered in the trajectory space and structurally reliable, MDS constructs a compact multi-turn training set that is simultaneously semantically diverse and well formed.

We validate MDS on two multi-turn training corpora, one general-purpose assistant dataset and one domain-specific customer-service dataset, each under a fixed 10K-dialogue selection budget. We compare against strong single-turn selection methods adapted to dialogue turns, various LLM-based multi-turn selectors, and several simple baselines. Across datasets and metrics, MDS consistently matches or surpasses these baselines under both reference-free and reference-based automatic evaluation, with particularly clear gains on measures of content coverage and fidelity. Beyond the main results, we conduct ablations that isolate the contribution of each component, and we provide in-depth analyses, visualizations, and case studies that show how MDS suppresses noisy conversations while preserving diverse yet well structured dialogues. Overall, our contributions are two-fold:
\begin{itemize}
    \item We propose MDS, a two-stage global and local framework for selecting multi-turn supervision based on semantic coverage and structural quality.
    \item We demonstrate that MDS consistently improves multi-turn performance over state-of-the-art selection and filtering schemes on both general-purpose and domain-specific corpora. We further introduce structural diagnostics that explain which types of dialogue-level noise MDS suppresses in practice. 
\end{itemize}
\section{Multi-turn Dialogue Selection}

\subsection{Problem Setup and Overview of MDS}

Let $D = \{d_1, \dots, d_N\}$ be a pool of multi-turn dialogues.  
Each dialogue $d$ is a sequence of user-assistant exchanges
\begin{equation}
    d = \{(Q_1, A_1), \dots, (Q_T, A_T)\},
\end{equation}

where $Q_t$ and $A_t$ denote the user query and assistant response at turn $t$.  
Given a fixed budget of $M$ dialogues, our goal is to select a subset $D^{*}$ that provides the most useful supervision for multi-turn instruction tuning. Most existing data selection methods score isolated instruction-response pairs, ignoring the multi-turn conversational structure across turns. We instead score whole dialogues and target a subset that both covers diverse user intents and consists of structurally well-formed conversations, so the model learns from coherent, informative trajectories rather than noisy or repetitive exchanges.

To this end, we propose \textbf{MDS} (Multi-turn Dialogue Selection), a two-stage framework that combines \emph{global semantic coverage} with \emph{local structural quality}. In the global stage, MDS represents each dialogue by a user-query trajectory embedding, partitions the trajectory space into semantic bins, and performs bin-wise coverage selection with redundancy control to retain conversations that are representative yet diverse. In the local stage, MDS assesses each candidate dialogue using complementary structure signals, including entity-grounded coherence and novelty, as well as query-answer form consistency, and then prioritizes dialogues with stronger structural reliability.

\subsection{Global Stage: Semantic Coverage over Dialogues}
\label{sec:global}

The global stage constructs a dialogue candidate pool that \emph{covers} diverse user intents and interaction patterns while controlling redundancy. 
Instead of scoring individual turns, we perform selection in a dialogue-level \emph{trajectory space} derived from the user side. 
Concretely, we build representations from user queries rather than assistant responses: queries provide a stable signal of the underlying intent and task type, whereas responses often contain stylistic noise, templated phrasing, or low-quality content that can distort semantic grouping. 
Query-based trajectory representations therefore offer a cleaner basis for coverage-aware selection.

\paragraph{Query-trajectory representation.}
Given a dialogue $d$ with $T$ user turns, we encode each user query $Q_t$ into an embedding $q_t \in \mathbb{R}^h$ using a sentence encoder, and aggregate them into a dialogue-level \emph{query-trajectory} embedding:
\begin{equation}
    v_d = \frac{1}{T} \sum_{t=1}^{T} q_t .
\end{equation}
Each $v_d$ summarizes the overall semantic trajectory of the user requests, so dialogues centered on similar tasks tend to be close in this space.

\paragraph{Bin-wise semantic coverage.}
A single global ranking over $\{v_d\}$ is prone to being dominated by high-frequency templates, which can reduce coverage of long-tail intents.
To mitigate this, we partition the trajectory space into $K$ semantic bins and enforce selection within each bin.
We cluster $\{v_d\}_{d \in D}$ (e.g., with K-means), obtaining bins $\{B_k\}_{k=1}^{K}$ with centroids
\begin{equation}
    c_k = \frac{1}{|B_k|} \sum_{d \in B_k} v_d .
\end{equation}

\paragraph{Bin-wise coverage with redundancy control.}
Even within the same semantic bin, many dialogues can be near-duplicates. 
We therefore perform bin-wise greedy selection with redundancy control, which implements a practical coverage-diversity trade-off.
Let $S_k$ denote the set of dialogues already selected from bin $B_k$.
For a candidate $d_i \in B_k$, we define its representativeness and redundancy as
\begin{equation}
    s_i = \mathrm{sim}(v_{d_i}, c_k), \quad
    r_i = \max_{d_j \in S_k} \mathrm{sim}(v_{d_i}, v_{d_j}),
\end{equation}
where $\mathrm{sim}(\cdot,\cdot)$ is cosine similarity.
Starting from $S_k=\emptyset$, we iteratively add the next dialogue by maximizing a greedy marginal objective:
\begin{equation}
    d^{\mathrm{next}}
    = \arg\max_{d_i \in B_k \setminus S_k}
    \Bigl( \lambda s_i - (1-\lambda) r_i \Bigr),
\end{equation}
with $\lambda \in [0,1]$ and we set $\lambda = 0.5$ in our experiments. 

\paragraph{Output.}
We use the global stage to construct a reduced candidate pool for efficient local scoring. Specifically, for each bin $B_k$ we run the above procedure and keep the top $\alpha$ fraction of selected dialogues, denoted $\tilde{B}_k \subseteq B_k$ with $|\tilde{B}_k|=\lceil \alpha |B_k| \rceil$ (we use $\alpha=0.5$), and output the global candidate pool $D^{\mathrm{global}}=\bigcup_{k=1}^{K}\tilde{B}_k$.
This candidate pool maintains broad semantic coverage with redundancy control, while substantially reducing the computational cost of the local-stage scorer.
In practice, the global stage keeps the top $\alpha$ fraction per bin to form $D^{\mathrm{global}}$ for efficiency, while the final per-bin budget $m_k$ is applied in the local stage.

\subsection{Local Stage: Structural Quality within Dialogues}
\label{sec:local}

The local stage refines the candidate pool $D^{\mathrm{global}}$ by assessing \emph{within-dialogue structural reliability}. 
While the global stage targets semantic coverage in the trajectory space, the local stage focuses on whether a dialogue provides \emph{usable multi-turn supervision}: it filters conversations that drift off-topic, collapse into repetition, or exhibit systematic query-answer form mismatches, and then performs budgeted selection within the semantic bins defined by the global stage.
All signals in this stage are computed in a reference-free manner using a lightweight instruction-tuned scorer, making the procedure efficient and model-agnostic.

\paragraph{Signal 1: Entity-grounded coherence and novelty.}
Intuitively, a good multi-turn dialogue should maintain \emph{topic grounding} (staying anchored to user-mentioned entities) while ensuring \emph{information progress} (introducing new, informative content rather than repeating earlier responses). 
For each turn $t$, we prompt the scorer to extract three entity sets: entities in the current answer $E^A_t$, entities mentioned in user queries up to turn $t$, denoted $E^Q_{\le t}$, and entities appearing in previous answers $E^A_{<t}$.
We then quantify two complementary aspects at each turn:
(i) \emph{anchoring}, encouraging answers to stay grounded in what the user is asking about, and
(ii) \emph{novelty}, rewarding answers that introduce informative new entities rather than repeating prior content.
Formally, we define the per-turn entity score
\begin{equation}
\mathrm{ent}_t
= \frac{\bigl|E^A_t \cap E^Q_{\le t}\bigr|}{\lvert E^A_t \rvert}
+ \frac{\bigl|E^A_t \setminus E^A_{<t}\bigr|}{\lvert E^A_t \rvert},
\end{equation}
when $\lvert E^A_t \rvert > 0$ (and set $\mathrm{ent}_t = 0$ otherwise).
Averaging over turns yields a dialogue-level structural score:
\begin{equation}
s_{\mathrm{entity}}(d) = \frac{1}{T} \sum_{t=1}^{T} \mathrm{ent}_t.
\end{equation}

\paragraph{Signal 2: Query-answer form consistency.}
Beyond topical grounding, multi-turn supervision also requires \emph{functional alignment}: responses should match the form implied by the query type, such as step-by-step procedures for troubleshooting requests, explicit comparisons for comparative queries, or concrete recommendations for advice queries.
For each turn $t$, we prompt the scorer with $(Q_t, A_t)$ (and minimal context) to rate how well the \emph{form} of $A_t$ satisfies the expected form of $Q_t$ on a three-point scale $c_t \in \{0,1,2\}$.
We then define the dialogue-level form-consistency score as
\begin{equation}
s_{\mathrm{form}}(d) = \frac{1}{T} \sum_{t=1}^{T} c_t.
\end{equation}

\paragraph{Bin-wise budgeted refinement.}
After computing $s_{\mathrm{entity}}(d)$ and $s_{\mathrm{form}}(d)$ for all $d \in D^{\mathrm{global}}$, we perform budgeted selection within bins using the candidate sets $\{\tilde{B}_k\}$ from the global stage.
We first apply form consistency as a necessary-condition filter:
\begin{equation}
S_k^{\mathrm{form}} = \{\, d \in \tilde{B}_k : s_{\mathrm{form}}(d) \ge \tau_{\mathrm{form}} \,\}.
\end{equation}
We then allocate a per-bin budget $m_k$ proportional to the original bin size $|B_k|$ under the overall budget $M$ (so that $\sum_k m_k = M$), and within each $S_k^{\mathrm{form}}$ select the top $m_k$ dialogues ranked by $s_{\mathrm{entity}}(d)$ (or all of them if $|S_k^{\mathrm{form}}| < m_k$).
The union of these bin-level subsets forms the final training set $D^{*}$.

\section{Experimental Setup}

\subsection{Training and Evaluation Datasets}

We evaluate MDS on both general-purpose and domain-specific multi-turn corpora. For training, we use \textbf{Baize}~\cite{xu2023baize} as a general assistant corpus and \textbf{Banking}~\footnote{https://huggingface.co/datasets/talkmap/banking-conversation-corpus} as a domain-specific customer-service corpus. For evaluation, we adopt three public benchmarks (\textbf{MT-Eval}~\cite{Kwan2024MTEvalAM}, \textbf{ConsistentChat}~\cite{chen2025consistentchat}, \textbf{TopDial}~\cite{wang2023target}) and a \textbf{Banking Test} of 1{,}000 held-out Banking dialogues that are never used for training. These test sets jointly cover open-ended assistance, consistency-sensitive exchanges, and task-oriented dialogues; detailed statistics are provided in the Appendix~\ref{app:data}.

\subsection{Evaluation Metrics}

We use three types of metrics to evaluate multi-turn dialogue quality. \textbf{Reference-free metrics.} Because many user queries are open-ended, we rely on strong LLM judges. Specifically, we adopt \textit{LLM-EVAL}~\cite{lin2023llm} and \textit{G-EVAL}~\cite{liu2023g} with GPT-4o as the judge, scoring each dialogue on several 0-10 dimensions (e.g., helpfulness, relevance, coherence). For each test set, we report the average score on each dimension and the average across dimensions; prompts and rubrics are provided in the Appendix~\ref{app:eval}. \textbf{Reference-based metrics.} We further report an \textit{Ent-F1} score, obtained by using GPT-4o to extract entities from reference and generated answers and computing F1 over entities aggregated across turns, which reflects how well the model covers key entities in the gold dialogue. We also use a \textit{Cos (cosine similarity)} score, defined as the cosine similarity between sentence-level embeddings of the reference and generated answers computed by a Sentence-Transformers encoder (the \emph{all-MiniLM-L6-v2} variant\footnote{https://huggingface.co/sentence-transformers/all-MiniLM-L6-v2}). \textbf{Aggregate comparison.} Since these metrics have different scales, we also report the \textit{Average Rank} of each method across all metrics as a scale-free summary, where lower is better.

\begin{table*}[ht]
\centering
\small
\setlength{\tabcolsep}{1.8mm}
\renewcommand{\arraystretch}{1.1}
\begin{tabular}{lccccccccccccc}
\toprule[1.5pt]
\multicolumn{1}{l|}{}                         & \multicolumn{4}{c|}{\textbf{MT-Eval}}                                                 & \multicolumn{4}{c|}{\textbf{ConsistentChat}}                                          & \multicolumn{4}{c|}{\textbf{TopDial}}                                                 & \multirow{2}{*}{\textbf{\begin{tabular}[c]{@{}c@{}}Avg.\\ Rank\end{tabular}}} \\ \cline{2-13}
\multicolumn{1}{l|}{}                         & \textbf{L-E}  & \textbf{G-E}  & \textbf{Ent-F1} & \multicolumn{1}{c|}{\textbf{Cos}}   & \textbf{L-E}  & \textbf{G-E}  & \textbf{Ent-F1} & \multicolumn{1}{c|}{\textbf{Cos}}   & \textbf{L-E}  & \textbf{G-E}  & \textbf{Ent-F1} & \multicolumn{1}{c|}{\textbf{Cos}}   &                                                                               \\ \toprule[1.5pt]
\multicolumn{14}{l}{\textsc{Backbone: LLaMA3-8B-Instruct}}                                                                                                                                                                                                                                                                                                                                                     \\ \toprule[1.5pt]
\multicolumn{1}{c|}{\textbf{Backbone}}        & 8.04          & 7.44          & 0.569           & \multicolumn{1}{c|}{0.831}          & 8.11          & 6.73          & 0.222           & \multicolumn{1}{c|}{\textbf{0.808}} & \textbf{7.12} & \textbf{6.68} & {\ul 0.158}     & \multicolumn{1}{c|}{\textbf{0.499}} & 5.00                                                                          \\
\multicolumn{1}{c|}{\textbf{All Data}}        & {\ul 8.09}    & 7.41          & 0.567           & \multicolumn{1}{c|}{0.845}          & 8.42          & {\ul 7.20}    & {\ul 0.310}  & \multicolumn{1}{c|}{0.794}          & 6.61          & 6.20          & 0.139           & \multicolumn{1}{c|}{0.409}          & 5.67                                                                          \\
\multicolumn{1}{c|}{\textbf{Random Data}}          & 8.00          & 7.41          & 0.558           & \multicolumn{1}{c|}{0.842}          & {\ul 8.46}    & {\ul 7.20}    & 0.306           & \multicolumn{1}{c|}{0.783}          & 6.58          & 6.20          & 0.140           & \multicolumn{1}{c|}{0.409}          & 6.92                                                                          \\
\multicolumn{1}{c|}{\textbf{SuperFiltering}}  & 8.08          & 7.44          & 0.568           & \multicolumn{1}{c|}{{\ul 0.846}}    & 8.38          & 7.11          & 0.301           & \multicolumn{1}{c|}{0.788}          & 6.98          & 6.27          & 0.156           & \multicolumn{1}{c|}{0.458}          & 4.92                                                                          \\
\multicolumn{1}{c|}{\textbf{Rethinking}}      & 8.01          & 7.43          & 0.568           & \multicolumn{1}{c|}{0.845}          & 8.41          & 7.18          & {\ul 0.310}  & \multicolumn{1}{c|}{0.789}          & 6.81          & 6.36          & 0.144           & \multicolumn{1}{c|}{0.426}          & 5.50                                                                          \\
\multicolumn{1}{c|}{\textbf{ZIP}}             & 8.06          & 7.42          & {\ul 0.570}     & \multicolumn{1}{c|}{0.845}          & 8.42          & 7.18          & 0.294           & \multicolumn{1}{c|}{0.781}          & 6.78          & 6.28          & 0.153           & \multicolumn{1}{c|}{0.433}          & 5.58                                                                          \\
\multicolumn{1}{c|}{\textbf{DialScore}}      & 8.05          & 7.21          & 0.567           & \multicolumn{1}{c|}{0.845}          & 8.44          & 7.13          & 0.307           & \multicolumn{1}{c|}{0.787}          & 6.90          & 6.29          & 0.145           & \multicolumn{1}{c|}{0.430}          & 6.00                                                                          \\
\multicolumn{1}{c|}{\textbf{Heuristic}} & 7.99          & 7.23          & 0.562           & \multicolumn{1}{c|}{0.838}          & 8.43          & 7.10          & 0.300           & \multicolumn{1}{c|}{0.792}          & 6.97          & 6.20          & 0.151           & \multicolumn{1}{c|}{0.436}          & 7.00                                                                          \\
\multicolumn{1}{c|}{\textbf{CC-Score}}              & 8.05          & {\ul 7.48}    & {\ul 0.570}     & \multicolumn{1}{c|}{0.845}          & 8.41          & 7.16          & 0.305           & \multicolumn{1}{c|}{0.788}          & 6.82          & 6.38          & 0.151           & \multicolumn{1}{c|}{0.436}          & {\ul 4.58}                                                                    \\
\multicolumn{1}{c|}{\textbf{MDS}}             & \textbf{8.16} & \textbf{7.52} & \textbf{0.584}  & \multicolumn{1}{c|}{\textbf{0.857}} & \textbf{8.52} & \textbf{7.26} & {\textbf{0.316}}     & \multicolumn{1}{c|}{{\ul 0.797}}    & \textbf{7.12} & {\ul 6.48}    & \textbf{0.173}  & \multicolumn{1}{c|}{{\ul 0.465}}    & \textbf{1.25}                                                                 \\ \toprule[1.5pt]
\multicolumn{14}{l}{\textsc{Backbone: Qwen3-8B-Instruct}}                                                                                                                                                                                                                                                                                                                                                      \\ \toprule[1.5pt]
\multicolumn{1}{c|}{\textbf{Backbone}}        & 7.81          & 7.90          & 0.496           & \multicolumn{1}{c|}{0.826}          & 6.68          & 7.11          & 0.184           & \multicolumn{1}{c|}{0.711}          & \textbf{7.71} & \textbf{8.25} & {\ul 0.145}     & \multicolumn{1}{c|}{0.392}          & 7.25                                                                          \\
\multicolumn{1}{c|}{\textbf{All Data}}        & 7.90          & 8.08          & 0.568           & \multicolumn{1}{c|}{0.843}          & 8.28          & 7.96          & 0.301           & \multicolumn{1}{c|}{0.793}          & 7.14          & 7.54          & 0.123           & \multicolumn{1}{c|}{0.390}          & 7.25                                                                          \\
\multicolumn{1}{c|}{\textbf{Random Data}}          & 7.96          & 8.05          & 0.558           & \multicolumn{1}{c|}{0.844}          & 8.31          & 7.98          & 0.312           & \multicolumn{1}{c|}{0.799}          & 7.15          & 7.52          & 0.134           & \multicolumn{1}{c|}{0.426}          & 5.58                                                                          \\
\multicolumn{1}{c|}{\textbf{SuperFiltering}}  & 8.01          & 8.20          & 0.581           & \multicolumn{1}{c|}{0.847}          & 8.26          & 7.96          & 0.310           & \multicolumn{1}{c|}{{\ul 0.802}}    & 7.16          & 7.41          & 0.117           & \multicolumn{1}{c|}{0.411}          & 5.33                                                                          \\
\multicolumn{1}{c|}{\textbf{Rethinking}}      & 7.91          & 8.08          & 0.575           & \multicolumn{1}{c|}{0.846}          & 8.29          & {\ul 8.00}    & 0.295           & \multicolumn{1}{c|}{0.792}          & 7.22          & 7.62          & 0.112           & \multicolumn{1}{c|}{0.423}          & 5.75                                                                          \\
\multicolumn{1}{c|}{\textbf{ZIP}}             & 7.98          & 8.12          & 0.564           & \multicolumn{1}{c|}{0.841}          & 8.32          & 7.98          & {\ul 0.314}     & \multicolumn{1}{c|}{0.791}          & 7.13          & 7.58          & 0.110           & \multicolumn{1}{c|}{0.414}          & 6.17                                                                          \\
\multicolumn{1}{c|}{\textbf{DialScore}}      & 7.92          & {\ul 8.21}    & {\ul 0.585}     & \multicolumn{1}{c|}{\textbf{0.850}} & 8.32          & 7.97          & 0.300           & \multicolumn{1}{c|}{{\ul 0.802}}    & 7.14          & 7.56          & 0.116           & \multicolumn{1}{c|}{0.421}          & 4.67                                                                          \\
\multicolumn{1}{c|}{\textbf{Heuristic}} & 8.00          & 8.10          & 0.553           & \multicolumn{1}{c|}{0.828}          & {\ul 8.36}    & {\ul 8.00}    & 0.307           & \multicolumn{1}{c|}{0.793}          & 7.21          & 7.55          & 0.114           & \multicolumn{1}{c|}{0.412}          & 5.67                                                                          \\
\multicolumn{1}{c|}{\textbf{CC-Score}}              & {\ul 8.05}    & 8.18          & 0.579           & \multicolumn{1}{c|}{0.845}          & 8.33          & 7.97          & 0.299           & \multicolumn{1}{c|}{0.798}          & 7.16          & 7.46          & 0.128           & \multicolumn{1}{c|}{{\ul 0.431}}    & {\ul 4.75}                                                                    \\
\multicolumn{1}{c|}{\textbf{MDS}}             & \textbf{8.16} & \textbf{8.26} & \textbf{0.593}  & \multicolumn{1}{c|}{{\ul 0.848}}    & \textbf{8.44} & \textbf{8.04} & \textbf{0.338}  & \multicolumn{1}{c|}{\textbf{0.822}} & {\ul 7.32}    & {\ul 7.70}    & \textbf{0.150}  & \multicolumn{1}{c|}{\textbf{0.451}} & \textbf{1.25}                                                                 \\ \toprule[1.5pt]
\end{tabular}
\caption{Main results of MDS and baseline selection methods on Baize dataset and three multi-turn benchmarks. \textbf{L-E} and \textbf{G-E} denote \textbf{LLM-EVAL} and \textbf{G-EVAL}, respectively; \textbf{Ent-F1} and \textbf{Cos} denote entity-level F1 and embedding cosine similarity, respectively. All reported scores are averaged over 5 runs for each method. Bold numbers denote the best score and underlined numbers denote the second-best score in each column; the rightmost column reports the average rank over all 12 metrics, where lower is better.}

\label{tab:main}
\end{table*}

\subsection{Baseline Methods}

We compare MDS against three groups of methods.

\textbf{1) Single-turn selection.}
We include three state-of-the-art selectors: \texttt{SuperFiltering}~\cite{li2024superfiltering}, \texttt{Rethinking}~\cite{xia2024rethinking}, and \texttt{ZIP}~\cite{Yin2024EntropyLT}. Since they operate on query-answer pairs, we adapt them to dialogues by scoring each turn, aggregating turn-level scores into a single dialogue score, and selecting dialogues under the same 10K-dialogue budget as MDS. 

\textbf{2) Multi-turn selection.}
We consider three dialogue-level selectors: (i) A consistency-focused scoring baseline from \texttt{ConsistentChat} \textbf{(CC-Score)}~\cite{chen2025consistentchat}. We use \texttt{Qwen3-32B} and their released prompts (without modification) to evaluate dialogue quality. (ii) A simple \textbf{DialScore} baseline that directly prompts the same model to assign a single 1--10 overall score to each dialogue under a generic rubric. (iii) A lightweight \textbf{Heuristic} baseline that filters dialogues by simple statistics (e.g., proportion of very short answers, self-repetition, lexical diversity) and rank by a composite heuristic score. Please refer to Appendix~\ref{sec:heuristic_filter} for more details.

\textbf{3) Others.}
We also report \texttt{Random Data} (uniformly sampling 10K dialogues), \texttt{All Data} (using all available training dialogues), and the unfine-tuned \texttt{Backbone}. These baselines help disentangle the effect of data selection from model capacity and training budget. For all selection methods, including MDS, we ultimately obtain a 10K-dialogue training subset from each corpus and expand it into turn-level supervision for fine-tuning.

\subsection{MDS Configuration}

We fine-tune \textbf{LLaMA3-8B-Instruct} and \textbf{Qwen3-8B-Instruct} with LoRA adapters ($r{=}64$, $\alpha{=}128$, dropout $0.1$) on each 10K-dialogue subset. We use batch\_size=2, gradient\_accumulation\_steps=16, num\_train\_epochs=3, learning\_rate=1e-5, and warmup\_ratio=0.05 with a cosine schedule. In the global stage, we encode user queries with a Sentence-Transformers encoder (\emph{all-MiniLM-L6-v2}) to obtain dialogue-level trajectory embeddings, cluster them into $K{=}1000$ semantic bins $\{B_k\}$ with K-means, keeping the top $50\%$ dialogues per bin to form the candidate pool $D^{\mathrm{global}}$. In the local stage, we use Qwen3-8B-Instruct with greedy decoding to compute both the entity coherence–novelty score $s_{\mathrm{entity}}(d)$ and the form-consistency score $s_{\mathrm{form}}(d)$, applying simple normalization for entities and a fixed threshold $\tau_{\mathrm{form}}{=}1.0$ on $s_{\mathrm{form}}(d)$ to filter out low-quality dialogues. The prompt is provided in the Appendix~\ref{app:local_prompt}. We allocate bin-level quotas $m_k$ proportional to $|B_k|$ with rounding so that $\sum_k m_k = 10{,}000$, and within each bin keep the top-$m_k$ dialogues ranked by $s_{\mathrm{entity}}(d)$ after the form filter, yielding the final subset $D^{*}$ for each corpus. Selection is performed entirely offline, and the resulting subsets are used for all backbones.

\section{Main Results}

\subsection{General-Domain Results}

Table~\ref{tab:main} shows that MDS delivers consistent gains across backbones, achieving the best average rank on both backbones, which indicates that the improvements are not model-specific. MDS also mitigates degradation from training on the full noisy pool on the task-oriented TopDial benchmark: for LLaMA3-8B, \texttt{All Data} reduces TopDial L-E from 7.12 to 6.61, while MDS preserves 7.12 and attains the best TopDial Ent-F1 (0.173). This pattern suggests that indiscriminate multi-turn supervision can be harmful, and coverage-aware selection helps retain task-relevant dialogue behaviors.

MDS yields the most consistent improvements on structure-sensitive signals, aligning with our goal of improving within-dialogue reliability. In contrast, adapting strong single-turn selectors to dialogues remains insufficient, and even dialogue-aware baselines lag behind MDS, highlighting the need to control both dialogue-level coverage and within-dialogue structure. Notably, these gaps persist across all three benchmarks, showing that MDS improves not only general helpfulness scores but also consistency-oriented measures that reflect multi-turn quality.

To check robustness, we re-scored all outputs with Qwen3-32B as the judge. The two evaluators agree on 92.1\% of \textbf{instance-level} pairwise preferences; at the \textbf{system level} (ranking methods by their average scores), their rankings are also highly correlated (Spearman's $\rho = 0.89$). This suggests our conclusions are not tied to a particular judge.

\begin{table}[h]
\centering
\small
\setlength{\tabcolsep}{2.3mm}
\renewcommand{\arraystretch}{1.1}
\begin{tabular}{c|cc|cc}
\toprule[1.5pt]
                        & \multicolumn{2}{c|}{\textbf{Banking Test}} & \multicolumn{2}{c}{\textbf{ConsistentChat}} \\ \cline{2-5} 
                        & \textbf{G-E}        & \textbf{Ent-F1}       & \textbf{G-E}         & \textbf{Ent-F1}       \\ \toprule[1.5pt]
\textbf{Backbone}       & 6.28                & 0.184                & 6.70                 & 0.222                \\
\textbf{All Data}       & 6.58                & \textbf{0.354}       & 7.12                 & 0.288                \\
\textbf{Random Data}         & 6.42                & 0.313                & 7.22                 & 0.290                \\ \hline
\textbf{SuperFiltering} & 6.44                & 0.305                & 7.12                 & 0.283                \\
\textbf{Rethinking}     & 6.62                & 0.333                & 7.18                 & 0.282                \\
\textbf{ZIP}            & 6.60                & 0.304                & 7.16                 & 0.285                \\
\textbf{DialScore}   & 6.50                & 0.321                & 7.20                 & {\ul 0.291}          \\
\textbf{Heuristic}      & 6.58                & 0.321                & {\ul 7.22}           & 0.278                \\
\textbf{CC-Score}             & {\ul 6.64}          & 0.319                & 7.16                 & 0.285                \\
\textbf{MDS}            & \textbf{6.72}       & {\ul 0.351}          & \textbf{7.30}        & \textbf{0.300}       \\ \toprule[1.5pt]
\end{tabular}
\caption{Domain-specific selection performances on the Banking corpus.}
\label{tab:banking}
\end{table}

\subsection{Domain-Specific Results}

Table~\ref{tab:banking} reports results when all methods select 10K dialogues from the Banking corpus and we evaluate both in-domain (Banking Test) and out-of-domain (ConsistentChat) performance. On Banking Test, MDS attains the highest G-E score (6.72) while nearly matching the best entity coverage (Ent-F1 0.351 vs.\ 0.354 for All Data), thus preserving the gains of using all dialogues but with better conversational quality. Out-of-domain on ConsistentChat, MDS again achieves the highest G-E and Ent-F1, indicating that selecting Banking dialogues via MDS does not simply overfit to the customer-service domain but yields supervision that transfers better to a different multi-turn benchmark. Overall, these results complement the general-domain findings in Table~\ref{tab:main} and show that MDS can enhance both in-domain robustness and cross-domain generalization for domain-specific dialogue pools.




\subsection{Ablation Study}

\begin{table}[h]
\small
\centering
\setlength{\tabcolsep}{2.3mm}
\renewcommand{\arraystretch}{1.1}
\begin{tabular}{c|ll|ll}
\toprule[1.5pt]
                         & \multicolumn{2}{c|}{\textbf{MT-Eval}}                                 & \multicolumn{2}{c}{\textbf{TopDial}}                                  \\ \cline{2-5} 
\textbf{}                & \multicolumn{1}{c}{\textbf{G-E}} & \multicolumn{1}{c|}{\textbf{Ent-F1}} & \multicolumn{1}{c}{\textbf{G-E}} & \multicolumn{1}{c}{\textbf{Ent-F1}} \\ \toprule[1.5pt]
\textbf{MDS}             & \textbf{7.52}                    & \textbf{0.584}                      & \textbf{6.48}                       & \textbf{0.173}                     \\
\textit{Global-only}     & 7.38                             & 0.580                               & 6.30                             & 0.157                              \\
\textit{Local-only}      & 7.44                             & 0.576                               & {\ul 6.46}                    & {\ul 0.162}                             \\
\textit{w/o Binning}     & 7.44                             & 0.570                               & 6.26                             & 0.145                        \\
\textit{w/o Form Filter} & 7.42                             & 0.574                               & {\ul 6.46}              & 0.148                              \\ \toprule[1.5pt]
\end{tabular}
\caption{Ablation on the components of MDS using LLaMA3-8B. All rows are variants of MDS.}
\label{tab:ablation}
\end{table}

Table~\ref{tab:ablation} reports ablations on the components of MDS.
Removing either stage degrades performance: the \emph{Global-only} variant and the \emph{Local-only} variant are consistently worse than \emph{Full MDS}, and neither matches its entity-level gains, showing that semantic coverage and structural scoring are complementary. Within the global stage, turning off binning (\emph{w/o Binning}) keeps a similar G-E score but noticeably harms Ent-F1 on TopDial, indicating that semantic bins are important for preserving long-tail intents while de-duplicating frequent patterns. Within the local stage, removing the query-answer form filter (\emph{w/o Form Filter}) slightly changes G-E but reduces Ent-F1 on TopDial, confirming that hard filtering on form consistency contributes to higher-quality supervision. Detailed ablation for different numbers of semantic bins $K$ are provided in Appendix~\ref{app:bin_ablation}.

\section{Analysis}

\begin{figure}[h]
    \centering
    \includegraphics[width=\linewidth]{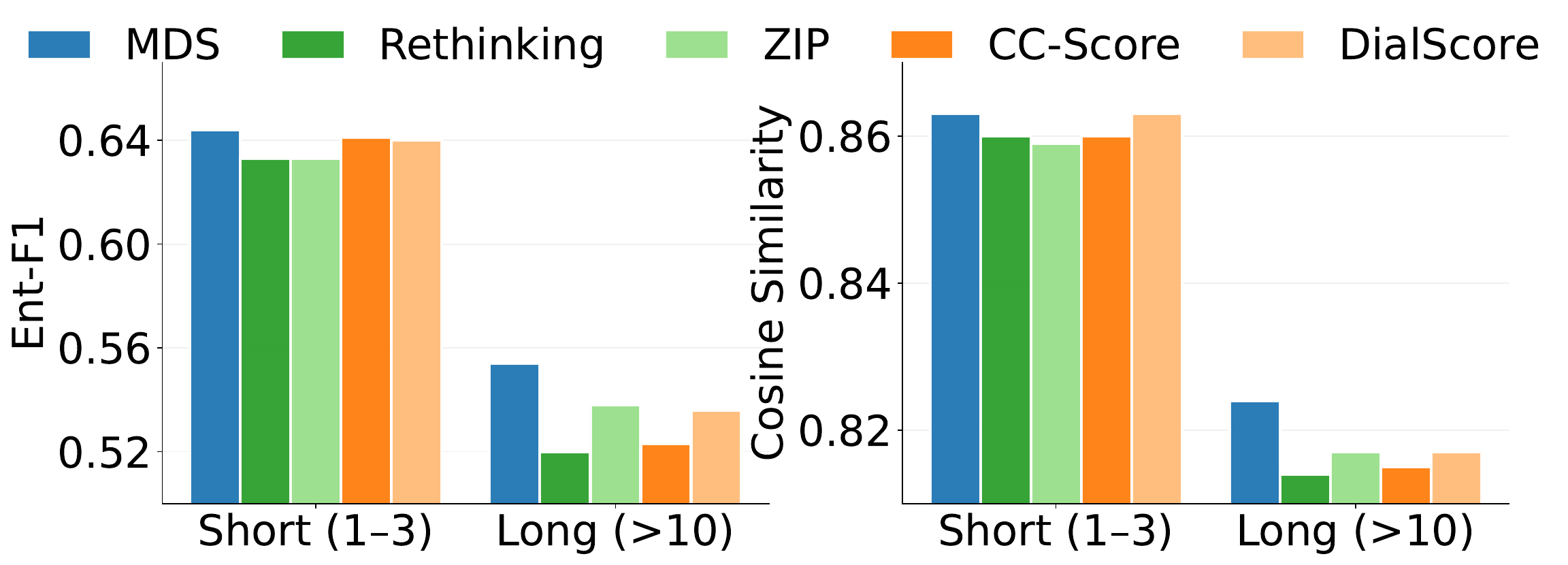}
    \caption{
    Performance on short (turns 1--3) vs.\ long (turns $>10$) queries on MT-Eval. We show Ent-F1 and cosine similarity averaged over turns in each bucket. Blue bars denote MDS, green bars adapted single-turn selectors (Rethinking, ZIP), and orange bars dialogue-level baselines (CC-Score, DialScore).
    }
    \label{fig:length-robustness}
\end{figure}

\subsection{Robustness across Dialogue Lengths}

To examine how selection strategies behave as conversations grow longer, we bucket MT-Eval turns by position into \emph{short} (user queries at turns 1--3) and \emph{long} (turns $>10$), and recompute metrics within each bucket. Figure~\ref{fig:length-robustness} reports Ent-F1 and cosine similarity for five selection methods. 

On short turns, all methods perform similarly; MDS is slightly ahead on Ent-F1 and cosine similarity. On long turns, the gap becomes more pronounced. All methods lose Ent-F1, but MDS degrades the least and maintains a clear margin over the best baseline (0.554 vs.\ 0.538 for ZIP and 0.536 for DialScore), and a similar trend holds for cosine similarity. This suggests that combining global semantic coverage with local structural filtering yields training data that better preserves entity coverage and semantic fidelity in later turns, making MDS more robust to length-induced degradation.




\begin{table}[h]
\centering
\small
\setlength{\tabcolsep}{2.6mm}
\renewcommand{\arraystretch}{1.1}
\begin{tabular}{lcccc}
\toprule[1.5pt]
\multicolumn{1}{c|}{}                    & \multicolumn{1}{c|}{\textbf{All Selected}} & \multicolumn{3}{c}{\textbf{Top 20\% by $H(d)$}} \\ \cline{2-5} 
\multicolumn{1}{c|}{}                    & \multicolumn{1}{c|}{$\mathrm{ESC}$} & $\mathrm{ESC}$  & $\mathrm{HAR}$ & $\mathrm{ENR}$ \\ \toprule[1.5pt]
\multicolumn{1}{l|}{\textbf{MDS}}        & \multicolumn{1}{c|}{0.599}        & 0.614         & 0.514         & 0.714        \\ \hline
\multicolumn{5}{l}{\textit{shuffle level}}                                                                                  \\ \hline
\multicolumn{1}{l|}{\emph{Pair}}       & \multicolumn{1}{c|}{0.596}        & 0.602         & 0.497         & 0.707        \\
\multicolumn{1}{l|}{\emph{Block(k=2)}} & \multicolumn{1}{c|}{0.596}        & 0.606         & 0.504         & 0.708        \\
\multicolumn{1}{l|}{\emph{Block(k=4)}} & \multicolumn{1}{c|}{0.596}        & 0.603         & 0.498         & 0.707        \\
\multicolumn{1}{l|}{\emph{Query-only}}      & \multicolumn{1}{c|}{0.547}        & 0.560         & 0.407         & 0.713        \\ \toprule[1.5pt]
\end{tabular}
\caption{Order-perturbation analysis on the same 10K dialogues selected by MDS. 
We report the turn-weighted Entity Sequence Consistency score (ESC) on the full set (\textbf{All Selected}), and additionally report ESC together with two interpretable components on the Top 20\% high-history-dependency subset ranked by $H(d)$: History Anchoring Rate (HAR) and Entity Novelty Rate (ENR).}
\label{tab:order_perturb_analysis_compact}
\end{table}

\subsection{Order Perturbation Analysis: Quantifying Cross-Turn Dependency}\label{sec:order}

We conduct a controlled counterfactual analysis to isolate whether the gains of MDS are truly driven by preserving cross-turn structure. Specifically, we fix the training set to the same 10K dialogues selected by MDS and only apply order-level perturbations to the dialogue organization: \emph{Pair shuffle} performs local swaps of adjacent QA pairs, \emph{Block shuffle}$(k{=}2/4)$ reorders turns in larger blocks with higher disruption for larger $k$, and \emph{Query-only shuffle} breaks query-answer correspondence as a stronger content-mismatch baseline. We additionally evaluate a \emph{high-history-dependency} subset (Top 20\% by $H(d)$), where $H(d)$ is a dialogue-level score computed from our turn-wise structural signals to quantify how strongly later turns depend on earlier turns, characterized by higher history anchoring and lower entity novelty (i.e., more reuse of previously introduced entities).

For evaluation, we use \textbf{ESC} (\emph{Entity Sequence Consistency}) as an order-sensitive overall score, and further decompose Top 20\% behavior into two interpretable factors: \textbf{HAR} (\emph{History Anchoring Rate}) and \textbf{ENR} (\emph{Entity Novelty Rate})\footnote{All metric definitions and their exact computation formulas are provided in Appendix~\ref{app:order_metrics}.}. The results show that order shuffles primarily degrade cross-turn consistency on the high-dependency subset, and the degradation is mainly driven by weakened history anchoring (HAR), while novelty (ENR) remains relatively stable. This pattern indicates that the main failure mode of order perturbations is breaking history anchoring rather than reducing entity novelty, reinforcing our design choice of explicitly modeling both anchoring and redundancy in the local stage.

\subsection{Error-type Analysis on Difference Sets}

\begin{figure}[h]
    \centering
    \includegraphics[width=\linewidth]{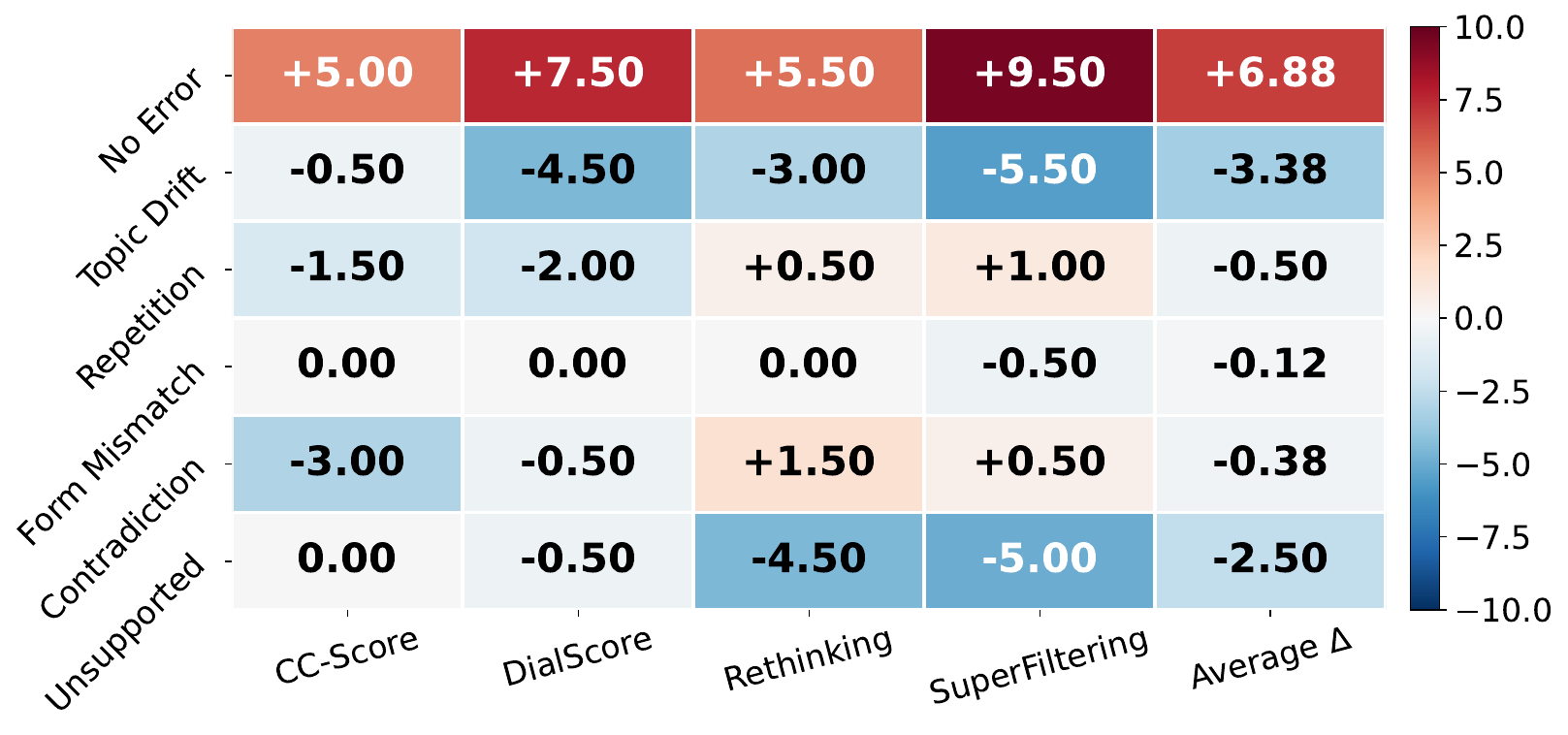}
    \caption{
    Error-type gaps on difference sets between MDS and each baseline selector. Each column compares MDS against one baseline. For example, in the \emph{DialScore} column, a cell value is the percentage-point gap \mbox{$\Delta = p(\emph{MDS-only}) - p(\emph{DialScore-only})$} for that error type. Detailed statistics are provided in the Appendix~\ref{app:error_type_supp}.
    }
    \label{fig:error-type-delta}
\end{figure}

To better understand what kinds of dialogues MDS prefers, we analyze \emph{difference sets} between MDS and strong baselines $B \in \{\text{CC-Score}, \text{DialScore}, \text{Rethinking}, \text{SuperFiltering}\}$. For each $B$, we construct \emph{MDS-only} ($\mathcal{D}_{\emph{MDS}}\setminus\mathcal{D}_{B}$) and \emph{Baseline-only} ($\mathcal{D}_{B}\setminus\mathcal{D}_{\emph{MDS}}$), uniformly sample 1{,}000 dialogues from each, and ask GPT-4o to assign a primary label from six categories: \textbf{No Error}, \textbf{Topic Drift}, \textbf{Repetition}, \textbf{Form Mismatch}, \textbf{Contradiction}, and \textbf{Unsupported}. We then compute the percentage-point gap $\Delta = p(\text{MDS-only}) - p(\text{Baseline-only})$ for each error type (Figure~\ref{fig:error-type-delta}), where negative $\Delta$ indicates fewer errors in MDS-only (while positive $\Delta$ is desirable for \textbf{No Error}). Appendix~\ref{app:error_type} shows the classification prompt. 

Figure~\ref{fig:error-type-delta} shows several consistent trends. MDS-only subsets contain noticeably more clean dialogues: \textbf{No Error} increases by $+5.0$ to $+9.5$ percentage points across baselines (avg.\ $+6.9$), indicating that MDS allocates more capacity to well-formed multi-turn supervision. At the same time, dialogue-level failures that directly hurt cross-turn learning are suppressed: \textbf{Topic Drift} is reduced for every baseline (avg.\ $-3.4$, up to $-5.5$), and \textbf{Unsupported} content also drops substantially (avg.\ $-2.5$, up to $-5.0$), suggesting that MDS-selected dialogues stay closer to the user’s intent and make fewer unjustified claims. By contrast, \textbf{Form Mismatch} is rare in both subsets, and gaps for \textbf{Repetition} and \textbf{Contradiction} are small and sometimes mixed, implying that they are not the main drivers of the observed gains. Overall, this error profile highlights MDS’s advantage: it reshapes the training pool toward on-topic, grounded, and structurally coherent conversations.

\section{Related Work}

A large body of work studies how to select high-quality supervision for single-turn instruction tuning, using carefully curated small datasets or automated selection based on LLM scores, weak-to-strong filtering, uncertainty or influence estimates, and distribution-matching objectives \cite{zhou2023lima,chen2024alpagasus,li2024quantity,li2024superfiltering,he2025fine,liu2024selectit,Zhang2025TheBI,xia2024less,wang2023self,Li2023OneSL,zhou2026fincardscardbasedanalystreranking,li2026instructiondataselectionanswer}. Recent work further shows that smaller models can act as selectors for larger models and that random selection can be a surprisingly strong baseline under controlled settings \cite{xia2024rethinking,mekala2024smaller}, but these approaches operate on isolated instruction--response pairs and do not model dialogue-level coverage or multi-turn structure.

In the multi-turn setting, prior work has mainly focused on constructing datasets and benchmarks rather than dialogue-level selectors. MT-Eval and related efforts evaluate multi-turn capabilities with GPT-based judges~\cite{Kwan2024MTEvalAM,liu2024what}, and corpora such as UltraChat, ConsistentChat, Baize, ShareGPT, LMSYS-Chat-1M, TopDial, and LIGHT provide large-scale synthetic or real conversations~\cite{ding2023enhancing,chen2025consistentchat,wang2023target,Urbanek2019LearningTS,xu2023baize}. However, their pipelines typically rely on rule-based cleaning or generic LLM filtering at the utterance/turn level, without explicitly enforcing \emph{semantic coverage} and \emph{structural quality} for complete dialogues under a \emph{fixed selection budget}. MDS fills this gap by enforcing \emph{global semantic coverage} and complementing it with \emph{local structural scoring} for reliability within each dialogue.

\section{Conclusion}
In this paper, we proposed MDS (Multi-turn Dialogue Selection), a dialogue-level data selection framework for multi-turn instruction tuning. MDS combines a global coverage stage that selects representative yet non-redundant dialogue trajectories with a local structure stage that measures entity-level coherence and query-answer form consistency. Experiments on Baize and a Banking corpus show that MDS outperforms strong single-turn selectors, dialogue-level LLM scorers, and heuristic baselines across both reference-free and reference-based metrics. Ablation and analysis further indicate that MDS is more robust on long conversations and reduces topic drift and unsupported claims, suggesting that dialogue-level structure is a powerful signal for curating cleaner and more reliable supervision for conversational models.

\section*{Limitations}
A main limitation of MDS is that it does not substantially reduce \textit{Contradiction} errors in our error-type analysis. We suspect this is partly because MDS deliberately retains longer and structurally richer dialogues, where cross-turn dependencies make subtle inconsistencies and implicit conflicts harder to avoid, even when topic grounding and form consistency are satisfied.

\bibliography{custom}

@article{Ouyang2022TrainingLM,
  title={Training language models to follow instructions with human feedback},
  author={Ouyang, Long and Wu, Jeffrey and Jiang, Xu and Almeida, Diogo and Wainwright, Carroll and Mishkin, Pamela and Zhang, Chong and Agarwal, Sandhini and Slama, Katarina and Ray, Alex and others},
  journal={Advances in neural information processing systems},
  volume={35},
  pages={27730--27744},
  year={2022}
}

@inproceedings{Wang2022SuperNaturalInstructionsGV,
  title={Super-NaturalInstructions: Generalization via Declarative Instructions on 1600+ NLP Tasks},
  author={Yizhong Wang and Swaroop Mishra and Pegah Alipoormolabashi and Yeganeh Kordi and Amirreza Mirzaei and Anjana Arunkumar and Arjun Ashok and Arut Selvan Dhanasekaran and Atharva Naik and David Stap and Eshaan Pathak and Giannis Karamanolakis and Haizhi Gary Lai and Ishan Purohit and Ishani Mondal and Jacob Anderson and Kirby Kuznia and Krima Doshi and Maitreya Patel and Kuntal Kumar Pal and M. Moradshahi and Mihir Parmar and Mirali Purohit and Neeraj Varshney and Phani Rohitha Kaza and Pulkit Verma and Ravsehaj Singh Puri and Rushang Karia and Shailaja Keyur Sampat and Savan Doshi and Siddhartha Mishra and Sujan Reddy and Sumanta Patro and Tanay Dixit and Xudong Shen and Chitta Baral and Yejin Choi and Noah A. Smith and Hannaneh Hajishirzi and Daniel Khashabi},
  booktitle={Conference on Empirical Methods in Natural Language Processing},
  year={2022},
  url={https://api.semanticscholar.org/CorpusID:253098274}
}

@inproceedings{Urbanek2019LearningTS,
  title={Learning to Speak and Act in a Fantasy Text Adventure Game},
  author={Jack Urbanek and Angela Fan and Siddharth Karamcheti and Saachi Jain and Samuel Humeau and Emily Dinan and Tim Rockt{\"a}schel and Douwe Kiela and Arthur Szlam and Jason Weston},
  booktitle={Conference on Empirical Methods in Natural Language Processing},
  year={2019},
  url={https://api.semanticscholar.org/CorpusID:71144630}
}

@inproceedings{wang2023target,
  title={Target-oriented proactive dialogue systems with personalization: Problem formulation and dataset curation},
  author={Wang, Jian and Cheng, Yi and Lin, Dongding and Leong, Chak and Li, Wenjie},
  booktitle={Proceedings of the 2023 Conference on Empirical Methods in Natural Language Processing},
  pages={1132--1143},
  year={2023}
}

@article{mekala2024smaller,
  title={Smaller Language Models are capable of selecting Instruction-Tuning Training Data for Larger Language Models},
  author={Dheeraj Mekala and Alex Nguyen and Jingbo Shang},
  journal={ArXiv},
  year={2024},
  volume={abs/2402.10430},
  url={https://api.semanticscholar.org/CorpusID:267740312}
}

@inproceedings{Li2023OneSL,
  title={One Shot Learning as Instruction Data Prospector for Large Language Models},
  author={Yunshui Li and Binyuan Hui and Xiaobo Xia and Jiaxi Yang and Min Yang and Lei Zhang and Shuzheng Si and Ling-Hao Chen and Junhao Liu and Tongliang Liu and Fei Huang and Yongbin Li},
  booktitle={Annual Meeting of the Association for Computational Linguistics},
  year={2023},
  url={https://api.semanticscholar.org/CorpusID:266348323}
}

@inproceedings{wang2023self,
  title={Self-Instruct: Aligning Language Models with Self-Generated Instructions},
  author={Yizhong Wang and Yeganeh Kordi and Swaroop Mishra and Alisa Liu and Noah A. Smith and Daniel Khashabi and Hannaneh Hajishirzi},
  booktitle={Annual Meeting of the Association for Computational Linguistics},
  year={2022},
  url={https://api.semanticscholar.org/CorpusID:254877310}
}

@article{xia2024rethinking,
  title={Rethinking data selection at scale: Random selection is almost all you need},
  author={Xia, Tingyu and Yu, Bowen and Dang, Kai and Yang, An and Wu, Yuan and Tian, Yuan and Chang, Yi and Lin, Junyang},
  journal={arXiv preprint arXiv:2410.09335},
  year={2024}
}

@article{Peng2023InstructionTW,
  title={Instruction Tuning with GPT-4},
  author={Baolin Peng and Chunyuan Li and Pengcheng He and Michel Galley and Jianfeng Gao},
  journal={ArXiv},
  year={2023},
  volume={abs/2304.03277},
  url={https://api.semanticscholar.org/CorpusID:257985497}
}

@article{xia2024less,
  title={Less: Selecting influential data for targeted instruction tuning},
  author={Xia, Mengzhou and Malladi, Sadhika and Gururangan, Suchin and Arora, Sanjeev and Chen, Danqi},
  journal={arXiv preprint arXiv:2402.04333},
  year={2024}
}

@article{Zhang2025TheBI,
  title={The Best Instruction-Tuning Data are Those That Fit},
  author={Dylan Zhang and Qirun Dai and Hao Peng},
  journal={ArXiv},
  year={2025},
  volume={abs/2502.04194},
  url={https://api.semanticscholar.org/CorpusID:276161183}
}

@article{liu2024selectit,
  title={SelectIT: Selective Instruction Tuning for LLMs via Uncertainty-Aware Self-Reflection},
  author={Liangxin Liu and Xuebo Liu and Derek F. Wong and Dongfang Li and Ziyi Wang and Baotian Hu and Min Zhang},
  journal={Advances in Neural Information Processing Systems 37},
  year={2024},
  url={https://api.semanticscholar.org/CorpusID:268032188}
}

@inproceedings{he2025fine,
  title={FiNE: Filtering and Improving Noisy Data Elaborately with Large Language Models},
  author={He, Junliang and Fan, Ziyue and Kuang, Shaohui and Xiaoqing, Li and Song, Kai and Zhou, Yaqian and Qiu, Xipeng},
  booktitle={Proceedings of the 2025 Conference of the Nations of the Americas Chapter of the Association for Computational Linguistics: Human Language Technologies (Volume 1: Long Papers)},
  pages={8686--8707},
  year={2025}
}

@article{li2024superfiltering,
  title={Superfiltering: Weak-to-strong data filtering for fast instruction-tuning},
  author={Li, Ming and Zhang, Yong and He, Shwai and Li, Zhitao and Zhao, Hongyu and Wang, Jianzong and Cheng, Ning and Zhou, Tianyi},
  journal={arXiv preprint arXiv:2402.00530},
  year={2024}
}

@article{Yin2024EntropyLT,
  title={Entropy Law: The Story Behind Data Compression and LLM Performance},
  author={Mingjia Yin and Chuhan Wu and Yufei Wang and Hao Wang and Wei Guo and Yasheng Wang and Yong Liu and Ruiming Tang and Defu Lian and Enhong Chen},
  journal={ArXiv},
  year={2024},
  volume={abs/2407.06645},
  url={https://api.semanticscholar.org/CorpusID:271064746}
}

@article{lin2023llm,
  title={Llm-eval: Unified multi-dimensional automatic evaluation for open-domain conversations with large language models},
  author={Lin, Yen-Ting and Chen, Yun-Nung},
  journal={arXiv preprint arXiv:2305.13711},
  year={2023}
}

@article{liu2023g,
  title={G-eval: NLG evaluation using gpt-4 with better human alignment},
  author={Liu, Yang and Iter, Dan and Xu, Yichong and Wang, Shuohang and Xu, Ruochen and Zhu, Chenguang},
  journal={arXiv preprint arXiv:2303.16634},
  year={2023}
}

@inproceedings{
liu2024what,
title={What Makes Good Data for Alignment? A Comprehensive Study of Automatic Data Selection in Instruction Tuning},
author={Wei Liu and Weihao Zeng and Keqing He and Yong Jiang and Junxian He},
booktitle={The Twelfth International Conference on Learning Representations},
year={2024},
url={https://openreview.net/forum?id=BTKAeLqLMw}
}

@article{Kwan2024MTEvalAM,
  title={MT-Eval: A Multi-Turn Capabilities Evaluation Benchmark for Large Language Models},
  author={Wai-Chung Kwan and Xingshan Zeng and Yuxin Jiang and Yufei Wang and Liangyou Li and Lifeng Shang and Xin Jiang and Qun Liu and Kam-Fai Wong},
  journal={ArXiv},
  year={2024},
  volume={abs/2401.16745},
  url={https://api.semanticscholar.org/CorpusID:267320495}
}

@article{zheng2023lmsys,
  title={Lmsys-chat-1m: A large-scale real-world llm conversation dataset},
  author={Zheng, Lianmin and Chiang, Wei-Lin and Sheng, Ying and Li, Tianle and Zhuang, Siyuan and Wu, Zhanghao and Zhuang, Yonghao and Li, Zhuohan and Lin, Zi and Xing, Eric P and others.},
  journal={arXiv preprint arXiv:2309.11998},
  year={2023}
}

@article{xu2023baize,
  title={Baize: An open-source chat model with parameter-efficient tuning on self-chat data},
  author={Xu, Canwen and Guo, Daya and Duan, Nan and McAuley, Julian},
  journal={arXiv preprint arXiv:2304.01196},
  year={2023}
}

@inproceedings{ding2023enhancing,
  title={Enhancing chat language models by scaling high-quality instructional conversations},
  author={Ding, Ning and Chen, Yulin and Xu, Bokai and Qin, Yujia and Hu, Shengding and Liu, Zhiyuan and Sun, Maosong and Zhou, Bowen},
  booktitle={Proceedings of the 2023 Conference on Empirical Methods in Natural Language Processing},
  pages={3029--3051},
  year={2023}
}

@inproceedings{chen2025consistentchat,
  title={ConsistentChat: Building Skeleton-Guided Consistent Multi-Turn Dialogues for Large Language Models from Scratch},
  author={Chen, Jiawei and Guan, Xinyan and Yuan, Qianhao and Guozhao, Mo and Zhou, Weixiang and Lu, Yaojie and Lin, Hongyu and He, Ben and Sun, Le and Han, Xianpei},
  booktitle={Proceedings of the 2025 Conference on Empirical Methods in Natural Language Processing},
  pages={8426--8452},
  year={2025}
}

@article{yang2025qwen3,
  title={Qwen3 technical report},
  author={Yang, An and Li, Anfeng and Yang, Baosong and Zhang, Beichen and Hui, Binyuan and Zheng, Bo and Yu, Bowen and Gao, Chang and Huang, Chengen and Lv, Chenxuand and others.},
  journal={arXiv preprint arXiv:2505.09388},
  year={2025}
}

@misc{dubey2024llama,
  title        = {The Llama 3 Herd of Models},
  author       = {Dubey, Abhimanyu and Jauhri, Abhinav and Pandey, Abhinav and Kadian, Abhishek and Al-Dahle, Ahmad and Letman, Aiesha and Mathur, Akhil and Schelten, Alan and Fan, Angela and Yang, Amy and others},
  year         = {2024},
  eprint       = {2407.21783},
  archivePrefix= {arXiv},
  primaryClass = {cs.CL}
}

@article{qi2023fine,
  title={Fine-tuning aligned language models compromises safety, even when users do not intend to!},
  author={Qi, Xiangyu and Zeng, Yi and Xie, Tinghao and Chen, Pin-Yu and Jia, Ruoxi and Mittal, Prateek and Henderson, Peter},
  journal={arXiv preprint arXiv:2310.03693},
  year={2023}
}

@article{wang2023openchat,
  title={Openchat: Advancing open-source language models with mixed-quality data},
  author={Wang, Guan and Cheng, Sijie and Zhan, Xianyuan and Li, Xiangang and Song, Sen and Liu, Yang},
  journal={arXiv preprint arXiv:2309.11235},
  year={2023}
}

@article{shen2024rethinking,
  title={Rethinking data selection for supervised fine-tuning},
  author={Shen, Ming},
  journal={arXiv preprint arXiv:2402.06094},
  year={2024}
}

@inproceedings{li2024quantity,
  title={From quantity to quality: Boosting llm performance with self-guided data selection for instruction tuning},
  author={Li, Ming and Zhang, Yong and Li, Zhitao and Chen, Jiuhai and Chen, Lichang and Cheng, Ning and Wang, Jianzong and Zhou, Tianyi and Xiao, Jing},
  booktitle={Proceedings of the 2024 Conference of the North American Chapter of the Association for Computational Linguistics: Human Language Technologies (Volume 1: Long Papers)},
  pages={7602--7635},
  year={2024}
}

@misc{li2026instructiondataselectionanswer,
      title={Instruction Data Selection via Answer Divergence}, 
      author={Bo Li and Mingda Wang and Shikun Zhang and Wei Ye},
      year={2026},
      eprint={2604.10448},
      archivePrefix={arXiv},
      primaryClass={cs.CL},
      url={https://arxiv.org/abs/2604.10448}, 
}

@misc{zhao2026generatingeffectivecottraces,
      title={Generating Effective CoT Traces for Mitigating Causal Hallucination}, 
      author={Yiheng Zhao and Jun Yan},
      year={2026},
      eprint={2604.12748},
      archivePrefix={arXiv},
      primaryClass={cs.CL},
      url={https://arxiv.org/abs/2604.12748}, 
}

@inproceedings{tian-etal-2025-grv,
    title = "{GRV}-{KBQA}: A Three-Stage Framework for Knowledge Base Question Answering with Decoupled Logical Structure, Semantic Grounding and Structure-Aware Validation",
    author = "Tian, Yuhang  and
      Yang, Pan  and
      Song, Dandan  and
      Wu, Zhijing  and
      Wang, Hao",
    editor = "Christodoulopoulos, Christos  and
      Chakraborty, Tanmoy  and
      Rose, Carolyn  and
      Peng, Violet",
    booktitle = "Findings of the Association for Computational Linguistics: EMNLP 2025",
    month = nov,
    year = "2025",
    address = "Suzhou, China",
    publisher = "Association for Computational Linguistics",
    url = "https://aclanthology.org/2025.findings-emnlp.141/",
    doi = "10.18653/v1/2025.findings-emnlp.141",
    pages = "2618--2632",
    ISBN = "979-8-89176-335-7"
}

@inproceedings{tian-etal-2025-compkbqa,
    title = "{C}omp{KBQA}: Component-wise Task Decomposition for Knowledge Base Question Answering",
    author = "Tian, Yuhang  and
      Song, Dandan  and
      Wu, Zhijing  and
      Yang, Pan  and
      Zhou, Changzhi  and
      Yang, Jun  and
      Wang, Hao  and
      Ma, Huipeng  and
      Li, Chenhao  and
      Zhang, Luan",
    editor = "Christodoulopoulos, Christos  and
      Chakraborty, Tanmoy  and
      Rose, Carolyn  and
      Peng, Violet",
    booktitle = "Proceedings of the 2025 Conference on Empirical Methods in Natural Language Processing",
    month = nov,
    year = "2025",
    address = "Suzhou, China",
    publisher = "Association for Computational Linguistics",
    url = "https://aclanthology.org/2025.emnlp-main.16/",
    doi = "10.18653/v1/2025.emnlp-main.16",
    pages = "293--309",
    ISBN = "979-8-89176-332-6"
}

@misc{zhou2026fincardscardbasedanalystreranking,
      title={FinCARDS: Card-Based Analyst Reranking for Financial Document Question Answering}, 
      author={Yixi Zhou and Fan Zhang and Yu Chen and Haipeng Zhang and Preslav Nakov and Zhuohan Xie},
      year={2026},
      eprint={2601.06992},
      archivePrefix={arXiv},
      primaryClass={cs.IR},
      url={https://arxiv.org/abs/2601.06992}, 
}

@inproceedings{
chen2024alpagasus,
title={AlpaGasus: Training a Better Alpaca with Fewer Data},
author={Lichang Chen and Shiyang Li and Jun Yan and Hai Wang and Kalpa Gunaratna and Vikas Yadav and Zheng Tang and Vijay Srinivasan and Tianyi Zhou and Heng Huang and Hongxia Jin},
booktitle={The Twelfth International Conference on Learning Representations},
year={2024},
url={https://openreview.net/forum?id=FdVXgSJhvz}
}

@inproceedings{dong2024abilities,
  title={How abilities in large language models are affected by supervised fine-tuning data composition},
  author={Dong, Guanting and Yuan, Hongyi and Lu, Keming and Li, Chengpeng and Xue, Mingfeng and Liu, Dayiheng and Wang, Wei and Yuan, Zheng and Zhou, Chang and Zhou, Jingren},
  booktitle={Proceedings of the 62nd Annual Meeting of the Association for Computational Linguistics (Volume 1: Long Papers)},
  pages={177--198},
  year={2024}
}

@article{zhou2023lima,
  title={Lima: Less is more for alignment},
  author={Zhou, Chunting and Liu, Pengfei and Xu, Puxin and Iyer, Srinivasan and Sun, Jiao and Mao, Yuning and Ma, Xuezhe and Efrat, Avia and Yu, Ping and Yu, Lili and others.},
  journal={Advances in Neural Information Processing Systems},
  volume={36},
  pages={55006--55021},
  year={2023}
}

@misc{taori2023stanford,
  title={Stanford alpaca: An instruction-following llama model},
  author={Taori, Rohan and Gulrajani, Ishaan and Zhang, Tianyi and Dubois, Yann and Li, Xuechen and Guestrin, Carlos and Liang, Percy and Hashimoto, Tatsunori B},
  year={2023},
  publisher={Stanford, CA, USA}
}

@article{kopf2023openassistant,
  title={Openassistant conversations-democratizing large language model alignment},
  author={K{\"o}pf, Andreas and Kilcher, Yannic and Von R{\"u}tte, Dimitri and Anagnostidis, Sotiris and Tam, Zhi Rui and Stevens, Keith and Barhoum, Abdullah and Nguyen, Duc and Stanley, Oliver and Nagyfi, Rich{\'a}rd and others.},
  journal={Advances in neural information processing systems},
  volume={36},
  pages={47669--47681},
  year={2023}
}

\clearpage
\appendix
\section*{Appendix}

\section{Dataset statistics.}\label{app:data}
Table~\ref{tab:data_stats} summarizes the dialogue counts and average dialogue length used in our experiments.
We conduct selection on two large multi-turn dialogue pools, \textsc{Baize} (54,456 dialogues; 3.95 turns on average) and \textsc{Banking} (66,948 dialogues; 5.01 turns on average).
For evaluation, we report results on three multi-turn benchmarks, \textsc{MT-Eval} (130 dialogues; 7.30 turns), \textsc{ConsistentChat} (1,000 dialogues; 7.73 turns), and \textsc{TopDial} (1,321 dialogues; 5.11 turns).
In addition, we include \textsc{Banking Test} (1,000 dialogues; 4.98 turns) to assess domain-specific generalization on the Banking setting.

\begin{table}[h]
\begin{tabular}{c|cc}
\hline
                        & \textbf{\#Dialogues} & \textbf{Avg.turn} \\ \hline
\textbf{Baize}          & 54,456               & 3.95              \\
\textbf{Banking}        & 66,948               & 5.01              \\ \hline
\textbf{MT-Eval}        & 130                  & 7.30              \\
\textbf{ConsistentChat} & 1,000                & 7.73              \\
\textbf{TopDial}        & 1,321                & 5.11              \\
\textbf{Banking Test}   & 1,000                & 4.98              \\ \hline
\end{tabular}
\caption{Statistics of the dialogue selection pools and evaluation benchmarks. \#Dialogues denotes the number of multi-turn dialogues in each dataset, and Avg.\ turn denotes the average number of user--assistant turns per dialogue.}
\label{tab:data_stats}
\end{table}

\section{Order-Perturbation Metrics: ESC, HAR, ENR, and $H(d)$}
\label{app:order_metrics}

This appendix defines the three metrics used in our \emph{Order Perturbation Analysis} (Section~\ref{sec:order}) and the history-dependency score $H(d)$ used to form the \textbf{Top 20\% by $H(d)$} subset. All metrics are computed from the same turn-wise entity annotations produced by our scoring pipeline (i.e., the extracted \texttt{q\_entities} and \texttt{a\_entities} per turn).

\paragraph{Notation.}
A dialogue $d$ contains $T$ user--assistant turns (QA pairs), indexed by $t\in\{1,\dots,T\}$.
For each turn $t$, let $E_t^{Q}$ and $E_t^{A}$ denote the entity sets extracted from the user query and the assistant answer, respectively (corresponding to \texttt{q\_entities} and \texttt{a\_entities} in our pipeline).
We define the \emph{history entity set} before turn $t$ as
\begin{equation}
C_t \;=\; \bigcup_{j=1}^{t-1} \left( E_j^{Q} \cup E_j^{A} \right).
\end{equation}
Intuitively, $C_t$ summarizes all entities that have been introduced in the dialogue context up to (but excluding) the current turn.

\paragraph{History Anchoring Rate (HAR).}
HAR measures how well the current answer \emph{anchors} to the previously established entity context.
For a dialogue $d$, we denote the turn-level anchoring score at turn $t$ by $\mathrm{HAR}_d(t)$.
We compute it using an F1-style overlap between the answer entities $E_t^{A}$ and the history entities $C_t$:
\begin{equation}
\mathrm{HAR}_d(t) \;=\;
\begin{cases}
\frac{2|E_t^{A}\cap C_t|}{|E_t^{A}|+|C_t|}, & \text{if } |E_t^{A}|+|C_t|>0,\\
0, & \text{otherwise.}
\end{cases}
\label{eq:har_turn}
\end{equation}
We then define the dialogue-level HAR as the average over turns:
\begin{equation}
\mathrm{HAR}(d) \;=\; \frac{1}{T}\sum_{t=1}^{T}\mathrm{HAR}_d(t).
\label{eq:har_dialog}
\end{equation}
\textbf{Interpretation:} higher HAR indicates stronger reuse/grounding to previously mentioned entities, hence stronger cross-turn anchoring.

\paragraph{Entity Novelty Rate (ENR).}
ENR measures how many entities in the current answer are \emph{new} with respect to the prior context.
For a dialogue $d$, we denote the turn-level novelty score at turn $t$ by $\mathrm{ENR}_d(t)$.
We compute it as the fraction of answer entities not seen in the history:
\begin{equation}
\mathrm{ENR}_d(t) \;=\;
\begin{cases}
\frac{|E_t^{A} \setminus C_t|}{|E_t^{A}|}, & \text{if } |E_t^{A}|>0,\\
0, & \text{otherwise.}
\end{cases}
\label{eq:enr_turn}
\end{equation}
The dialogue-level ENR is again the average over turns:
\begin{equation}
\mathrm{ENR}(d) \;=\; \frac{1}{T}\sum_{t=1}^{T}\mathrm{ENR}_d(t).
\label{eq:enr_dialog}
\end{equation}
\textbf{Interpretation:} higher ENR indicates the answer introduces more new entities (less redundancy); lower ENR indicates the dialogue is more history-dependent, with heavier reuse of previously established entities.

\paragraph{Entity Sequence Consistency (ESC).}
ESC is an order-sensitive overall score that combines the above two complementary factors:
\begin{equation}
\mathrm{ESC}(d) \;=\; \frac{1}{2}\Big(\mathrm{HAR}(d) + \mathrm{ENR}(d)\Big).
\label{eq:esc_dialog}
\end{equation}
\textbf{Interpretation:} ESC is high when a dialogue simultaneously maintains strong history anchoring (HAR) while still introducing non-trivial new entities (ENR), which matches our local-stage design goal of balancing \emph{anchoring} and \emph{anti-redundancy}.

\paragraph{Turn-weighted aggregation (reported in Table~\ref{tab:order_perturb_analysis_compact}).}
For a dialogue set $\mathcal{D}$, let $T_d$ denote the number of turns in dialogue $d$.
We report turn-weighted scores so that each turn contributes equally:
\begin{equation}
\mathrm{HAR}_{\mathrm{tw}}(\mathcal{D}) \;=\; \frac{\sum_{d\in\mathcal{D}}\sum_{t=1}^{T_d}\mathrm{HAR}_d(t)}{\sum_{d\in\mathcal{D}}T_d},
\end{equation}
\begin{equation}
\mathrm{ENR}_{\mathrm{tw}}(\mathcal{D}) \;=\; \frac{\sum_{d\in\mathcal{D}}\sum_{t=1}^{T_d}\mathrm{ENR}_d(t)}{\sum_{d\in\mathcal{D}}T_d},
\label{eq:turn_weighted}
\end{equation}
and
\begin{equation}
\mathrm{ESC}_{\mathrm{tw}}(\mathcal{D}) \;=\; \frac{1}{2}\Big(\mathrm{HAR}_{\mathrm{tw}}(\mathcal{D}) + \mathrm{ENR}_{\mathrm{tw}}(\mathcal{D})\Big).
\label{eq:esc_turn_weighted}
\end{equation}

\paragraph{History-dependency score $H(d)$ for the Top-20\% subset.}
To focus on dialogues that require stronger cross-turn dependency, we compute a dialogue-level history-dependency score that increases with stronger anchoring and decreases with higher novelty:
\begin{equation}
H(d) \;=\; \tfrac{1}{2}\Big(\mathrm{HAR}(d) + \big(1-\mathrm{ENR}(d)\big)\Big).
\label{eq:Hd}
\end{equation}
We rank the 10K selected dialogues by $H(d)$ and take the top 20\% as the \emph{high-history-dependency} subset. The subset size can be slightly different from exactly 20\% due to ties in $H(d)$.

\paragraph{Why these metrics are order-sensitive.}
All four quantities above depend on the \emph{history set} $C_t$, which is defined by the turn order. Therefore, order-level perturbations (Pair/Block shuffles) alter $C_t$ for many turns and can reduce HAR/ESC even when the multiset of turns is unchanged. In contrast, \emph{Query-only} perturbation additionally breaks query--answer correspondence, yielding a stronger mismatch that typically collapses HAR.

\begin{figure}[h]
    \centering
    \includegraphics[width=\linewidth]{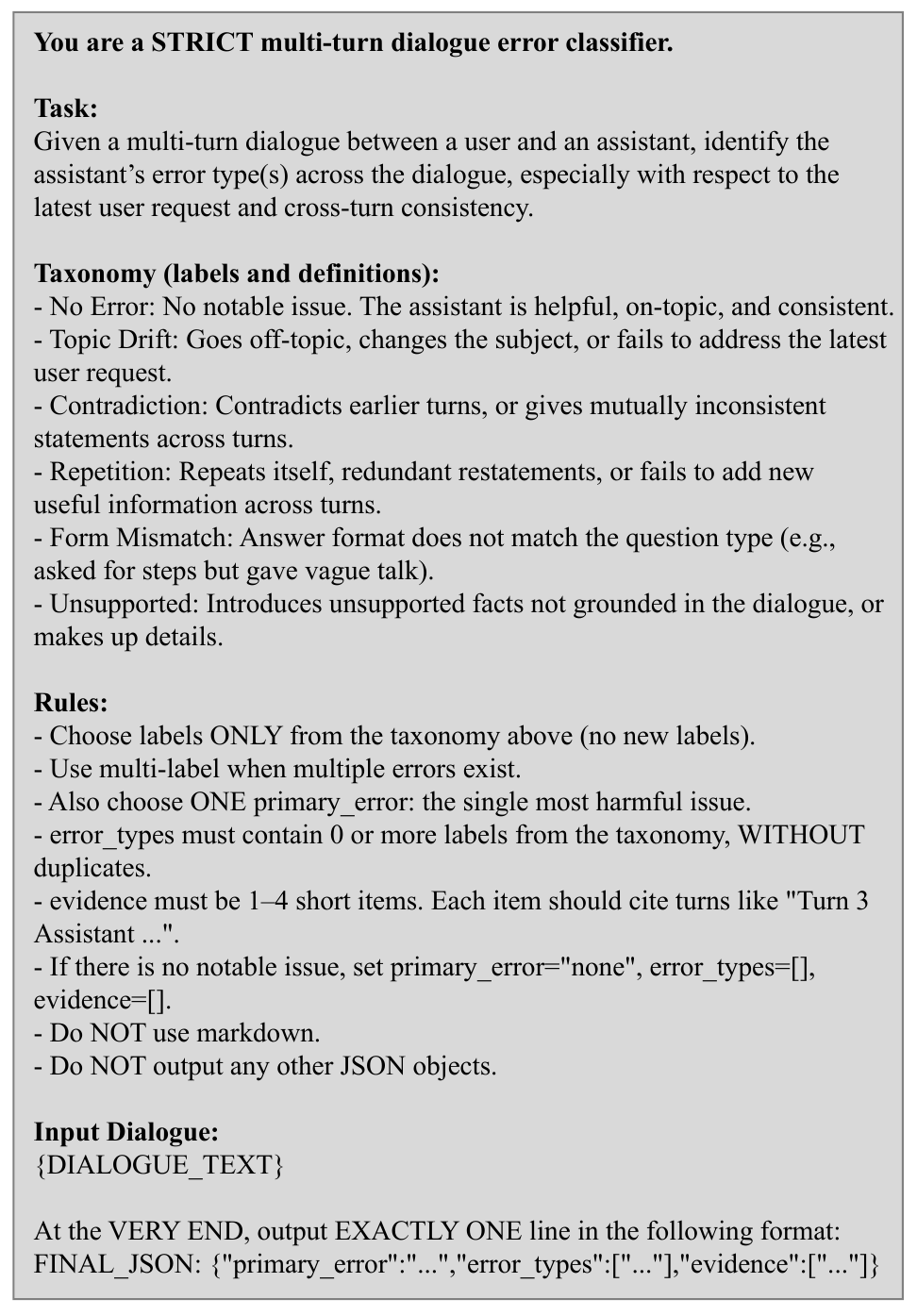}
    \caption{Prompt used for GPT-based multi-turn dialogue error-type classification. The judge assigns a primary error label and an optional set of additional error labels from a fixed taxonomy, and returns brief evidence by referencing specific turns.}
    \label{fig:prompt_error_type}
\end{figure}

\section{Error-Type Classifier Prompt}
\label{app:error_type}

To better understand the qualitative differences between dialogues selected by MDS and competing selectors, we perform an error-type analysis using GPT-4o as a strict multi-turn dialogue judge. 
Given a dialogue transcript, the judge is instructed to classify assistant-side failures using a fixed taxonomy that covers common multi-turn issues, including topic drift, contradiction/inconsistency, repetition/low novelty, form mismatch,  and unsupported/hallucinated content, with an additional \texttt{No Error} label indicating no notable issue. 
The judge must output a single JSON line containing (i) a \texttt{primary\_error} as the most harmful issue, (ii) an optional multi-label set \texttt{error\_types} without duplicates, and (iii) short \texttt{evidence} snippets that cite the relevant turns (e.g., ``Turn 3 Assistant \ldots''). 
This constrained format ensures consistent labeling across methods and enables reliable aggregation of error distributions for comparison.

\begin{table*}[t]
\centering
\small
\setlength{\tabcolsep}{2mm}
\renewcommand{\arraystretch}{1.15}
\begin{tabular}{l|cc|cc|cc|cc}
\toprule
\multirow{2}{*}{\textbf{Error type}} 
& \multicolumn{2}{c|}{\textbf{CC}}
& \multicolumn{2}{c|}{\textbf{DialScore}}
& \multicolumn{2}{c|}{\textbf{Rethinking}}
& \multicolumn{2}{c}{\textbf{SuperFiltering}} \\
& \textbf{MDS-only} & \textbf{B-only}
& \textbf{MDS-only} & \textbf{B-only}
& \textbf{MDS-only} & \textbf{B-only}
& \textbf{MDS-only} & \textbf{B-only} \\
\midrule
No Error                    & 87.5 & 82.5 & 87.0 & 79.5 & 86.0 & 80.5 & 86.5 & 77.0 \\
Topic Drift    &  9.5 & 10.0 &  5.0 &  9.5 &  6.0 &  9.0 &  6.0 & 11.5 \\
Repetition &  1.0 &  2.5 &  0.5 &  2.5 &  3.0 &  2.5 &  3.0 &  2.0 \\
Form Mismatch           &  0.0 &  0.0 &  0.0 &  0.0 &  0.0 &  0.0 &  0.0 &  0.5 \\
Contradiction 
                        &  1.0 &  4.0 &  4.5 &  5.0 &  3.5 &  2.0 &  2.5 &  2.0 \\
Unsupported              &  1.0 &  1.0 &  3.0 &  3.5 &  1.5 &  6.0 &  2.0 &  7.0 \\
\bottomrule
\end{tabular}
\caption{Error-type distribution on \textsc{difference sets}. For each baseline selector $B$, we compare 1K dialogues sampled from \textsc{MDS-only} ($\mathcal{D}_{\textsc{MDS}}\setminus \mathcal{D}_{B}$) versus \textsc{B-only} ($\mathcal{D}_{B}\setminus \mathcal{D}_{\textsc{MDS}}$). Values are percentages; higher \texttt{No Error} indicates cleaner dialogues.}
\label{tab:error_type_merged}
\end{table*}

\section{Supplementary Error-Type Analysis on Difference Sets}
\label{app:error_type_supp}

This section provides supplementary evidence to the main text by characterizing \emph{what kinds of dialogues} are uniquely favored by \textbf{MDS} compared to alternative selection methods. 
Rather than analyzing the full selected sets (which often share a large overlap), we follow a \emph{difference-set} protocol that isolates the distinctive portion of each selector.

\paragraph{Difference sets.}
For each baseline selector $B$, we construct two disjoint sets:
(i) \textbf{MDS-only}, $\mathcal{D}_{\textbf{MDS}}\setminus \mathcal{D}_B$, containing dialogues selected by \textbf{MDS} but not by $B$; and
(ii) \textbf{B-only}, $\mathcal{D}_B\setminus \mathcal{D}_{\textbf{MDS}}$, containing dialogues selected by $B$ but not by \textbf{MDS}.
This comparison controls for the shared subset and highlights the structural differences induced by the selection strategy.

\paragraph{Sampling and labeling.}
From each difference set, we uniformly sample 1,000 dialogues and assign each dialogue to one error type using the same taxonomy and the same LLM-based classifier described in Appendix~\ref{app:error_type}. 
The reported numbers are the percentages of dialogues in each error category.
A higher \textit{No Error} rate indicates cleaner and more coherent dialogues, while higher rates of others indicate specific failure modes.

\paragraph{Results overview.}
Table~\ref{tab:error_type_merged} summarizes the error-type distributions for four baselines.
Across baselines, \textbf{MDS-only} dialogues consistently exhibit a higher \textit{No Error} proportion and reduced rates of major multi-turn failure types, suggesting that \textbf{MDS} preferentially keeps dialogues with better cross-turn coherence and fewer structural issues.
These findings complement the main results by providing a data-level explanation of why \textbf{MDS}-selected dialogues lead to stronger downstream behavior.

\begin{figure}[h]
    \centering
    \includegraphics[width=\linewidth]{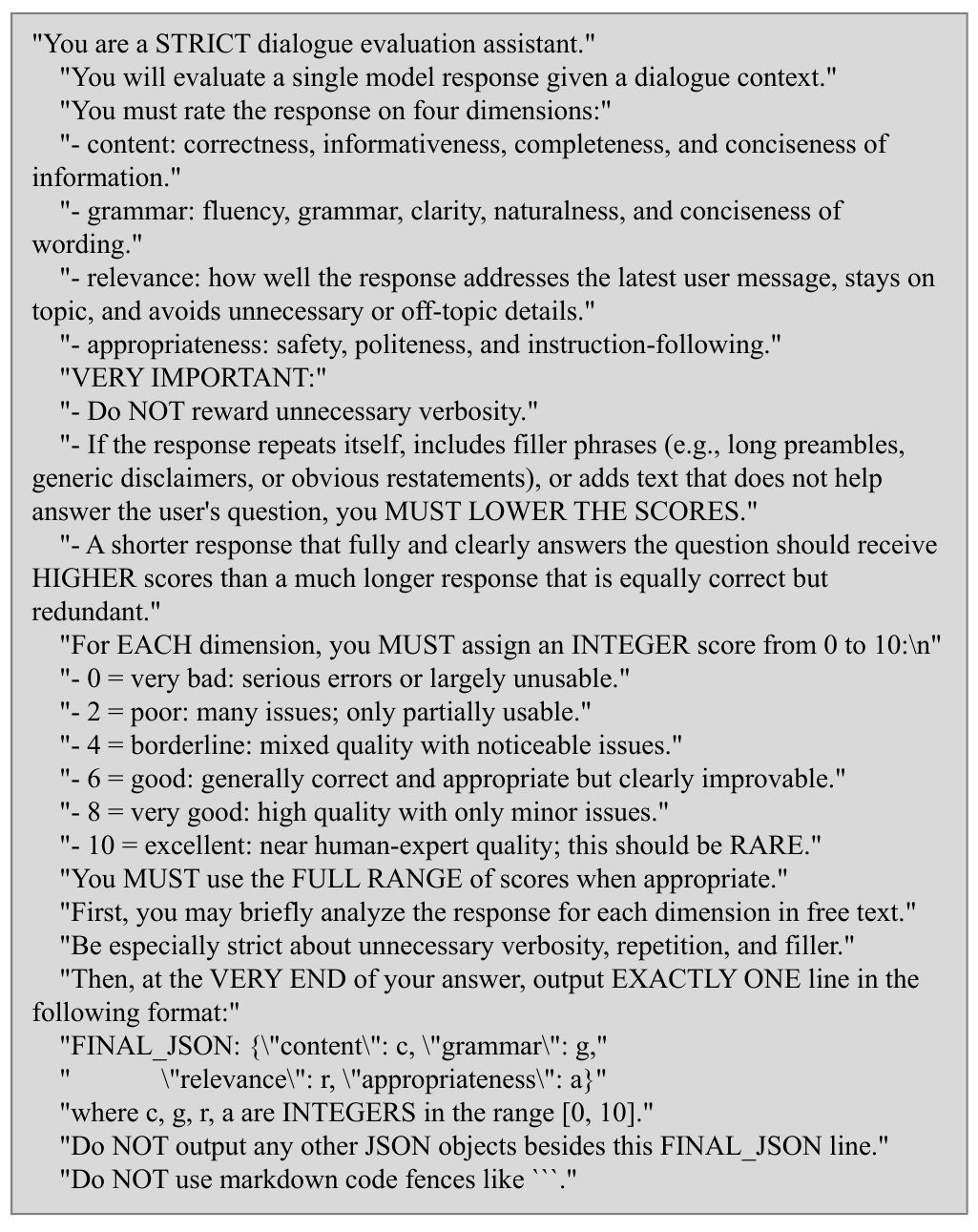}
    \caption{Prompt used for LLM-EVAL.}
    \label{fig:prompt_llm_eval}
\end{figure}

\begin{figure}[h]
    \centering
    \includegraphics[width=\linewidth]{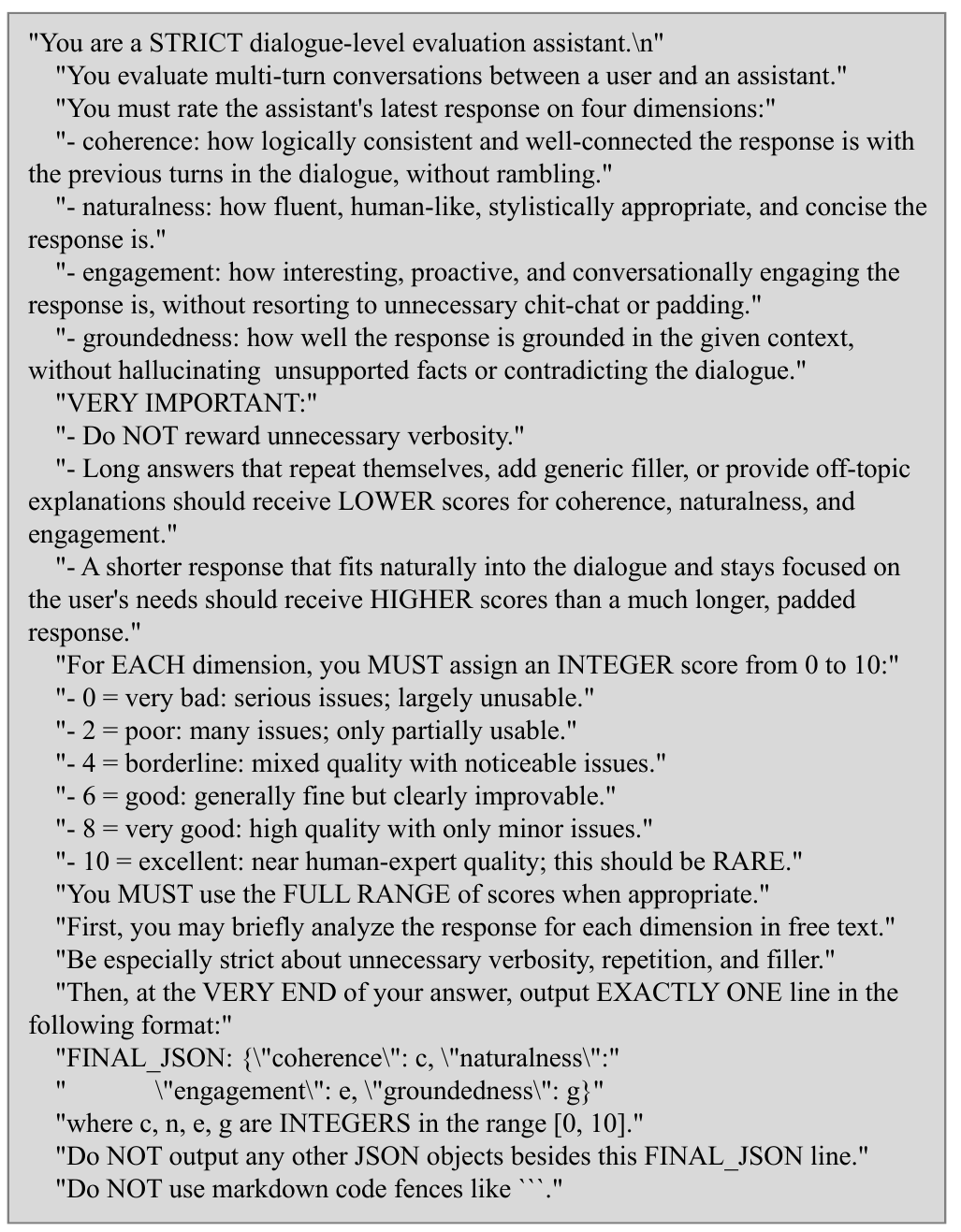}
    \caption{Prompt used for G-EVAL.}
    \label{fig:prompt_g_eval}
\end{figure}

\section{Prompt Used for LLM-EVAL and G-EVAL}
\label{app:eval}

Figure~\ref{fig:prompt_llm_eval} and Figure~\ref{fig:prompt_g_eval} show the prompt used for LLM-EVAL and G-EVAL. The above two metrics are both reference-free and rely on GPT-4o as the judge.

\begin{table*}[]
\centering
\small
\setlength{\tabcolsep}{1.8mm}
\renewcommand{\arraystretch}{1.1}
\begin{tabular}{c|cccc|cccc|cccc}
\hline
\multicolumn{1}{l|}{}  & \multicolumn{4}{c|}{\textbf{MT-Eval}}                            & \multicolumn{4}{c|}{\textbf{ConsistentChat}}                     & \multicolumn{4}{c}{\textbf{TopDial}}                             \\ \cline{2-13} 
\multicolumn{1}{l|}{}  & \textbf{L-E}  & \textbf{G-E}  & \textbf{Ent-F1} & \textbf{Cos}   & \textbf{L-E}  & \textbf{G-E}  & \textbf{Ent-F1} & \textbf{Cos}   & \textbf{L-E}  & \textbf{G-E}  & \textbf{Ent-F1} & \textbf{Cos}   \\ \hline
\textbf{100}           & 8.10          & 7.46          & 0.561           & 0.847          & 8.46          & 7.16          & 0.310           & 0.791          & 6.94          & 6.40          & 0.156           & 0.453          \\
\textbf{500}           & 8.08          & 7.48          & 0.568           & 0.843          & 8.48          & 7.20          & 0.305           & 0.792          & 6.84          & 6.38          & 0.145           & 0.440          \\
\textbf{1000(default)} & \textbf{8.16} & \textbf{7.52} & \textbf{0.584}  & \textbf{0.857} & \textbf{8.52} & \textbf{7.26} & \textbf{0.316}  & 0.797          & \textbf{7.12} & \textbf{6.48} & \textbf{0.173}  & \textbf{0.465} \\
\textbf{1500}          & 8.10          & 7.44          & 0.561           & 0.844          & 8.48          & 7.16          & 0.302           & 0.795          & 6.94          & 6.40          & 0.162           & 0.457          \\
\textbf{2000}          & 8.04          & 7.46          & 0.576           & 0.846          & 8.46          & 7.16          & 0.307           & \textbf{0.799} & 6.88          & 6.40          & 0.161           & 0.450          \\ \hline
\end{tabular}
\caption{Ablation on bin granularity on \textbf{Baize} with \texttt{LLaMA3-8B-Instruct}. We report performance under four metrics (L-E, G-E, Ent-F1, and Cos), where higher is better for all metrics. Bold numbers indicate the best score in each column.}
\label{tab:bin_ablation}
\end{table*}

\section{Ablation on Bin Granularity}
\label{app:bin_ablation}

\paragraph{Setup.}
We study the effect of bin granularity in our bin-wise selection pipeline on \textbf{Baize}, using LLaMA3-8B-Instruct as the backbone. We vary the number of bins $K\in\{100,500,1000,1500,2000\}$ and keep the total selection budget as well as all other settings unchanged. Table~\ref{tab:bin_ablation} reports results on MT-Eval, ConsistentChat, and TopDial.

\paragraph{Results and analysis.}
We observe a clear sweet spot at $K{=}1000$ (our default), which achieves the best performance on almost all metrics across all three benchmarks.
When $K$ is too small (e.g., 100 or 500), bins become overly coarse and mix heterogeneous dialogues, which weakens within-bin normalization and makes the final selection more sensitive to superficial biases.
In contrast, when $K$ is too large (e.g., 1500 or 2000), bins become fragmented and sparse, leading to unstable within-bin statistics and noisier quota allocation, which hurts overall quality.
A minor exception is the Cos score on \textsc{ConsistentChat}, where $K{=}2000$ is slightly higher; however, this does not translate into consistent gains on L-E/G-E/Ent-F1 or on other benchmarks.
Overall, these results support using a moderate bin granularity to balance within-bin comparability and statistical stability.

\begin{figure}[h]
    \centering
    \includegraphics[width=\linewidth]{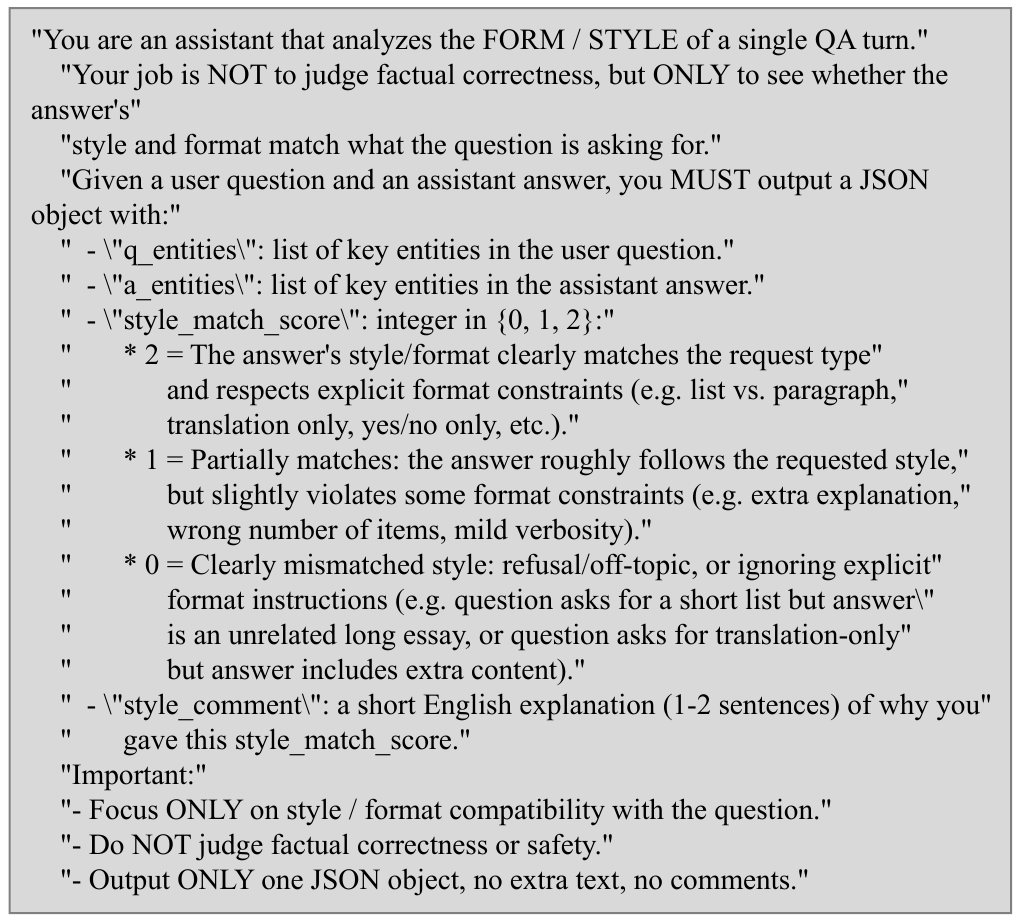}
    \caption{Local-stage structured scoring prompt used in MDS. We query Qwen3-8B with a single-turn QA pair and require a \texttt{JSON}-only output that \emph{simultaneously} extracts question/answer entities (\texttt{q\_entities}, \texttt{a\_entities}) and predicts a discrete form/style compatibility score (\texttt{style\_match\_score} $\in\{0,1,2\}$) with a brief rationale. This one-pass, multi-signal design enables efficient local scoring by avoiding separate calls for entity statistics and form assessment.}
    \label{fig:prompt_mds}
\end{figure}

\section{Local-stage Prompt for Joint Entity and Form/Style Scoring}
\label{app:local_prompt}

To compute local structural signals efficiently, we use a single structured prompt to obtain both (i) entity statistics and (ii) form/style compatibility judgments for each QA turn. Concretely, given a user question and the corresponding assistant answer, we query a lightweight judge model (Qwen3-8B) and require it to output \emph{only} one JSON object containing: (1) entities in the question (\texttt{q\_entities}) and in the answer (\texttt{a\_entities}), and (2) a discrete style-match score \texttt{style\_match\_score} $\in \{0,1,2\}$ indicating whether the answer's format matches what the question requests (e.g., list vs.\ paragraph, translation-only, yes/no-only), along with a short explanation \texttt{style\_comment}. 

This joint-output design is critical for efficiency: entity overlap statistics (used by our entity-based local signal) and form/style judgments (used by our form-based local signal) are produced in a \emph{single} forward pass per turn, rather than two separate model calls. As a result, local scoring can scale to large candidate pools with substantially reduced inference overhead while keeping the signals consistent by construction (both derived from the same model output and the same turn context).

\section{Heuristic Rule-based Dialogue Filtering}
\label{sec:heuristic_filter}

To construct a strong rule-based baseline for dialogue selection, we implement a lightweight heuristic filter that removes low-quality conversations and then ranks the remaining ones by a composite quality score. The filter operates on each dialogue independently and only uses surface statistics computed from the \emph{assistant} turns.

\paragraph{Preprocessing.}
For a dialogue $d$, we extract all assistant messages $\{a_1,\dots,a_T\}$ (with role normalized to \texttt{assistant}). Each message is tokenized into a word list using a Unicode-aware regex, and additionally split into sentences using punctuation-based segmentation. Dialogues with fewer than $\texttt{MIN\_ASST\_TURNS}$ assistant turns are discarded.

\paragraph{Quality constraints.}
We enforce three hard constraints to eliminate obvious noise:
\begin{itemize}
    \item \textbf{Short-response ratio.} We count an assistant turn as \emph{short} if its token length is below $\texttt{SHORT\_TOK\_TH}$ or its character length is below $\texttt{SHORT\_CHAR\_TH}$. Let $r_{\text{short}}$ be the fraction of short assistant turns. We discard $d$ if $r_{\text{short}} > \texttt{MAX\_SHORT\_RATIO}$.
    \item \textbf{Repetition score.} We measure repetition from both token-level and sentence-level perspectives. 
    First, we compute the $n$-gram repetition ratio using $n=\texttt{REP\_N}$ over the concatenated assistant token stream: 
    \[
        r_{n\text{g}} = 1 - \frac{|\text{unique }n\text{-grams}|}{|\text{all }n\text{-grams}|}.
    \]
    Second, we compute the duplicated-sentence ratio $r_{\text{sent}}$ as the fraction of assistant sentences that are exact repeats within the dialogue. We define the overall repetition score as
    \[
        r_{\text{rep}} = 0.5\, r_{n\text{g}} + 0.5\, r_{\text{sent}}.
    \]
    We discard $d$ if $r_{\text{rep}} > \texttt{MAX\_REP\_SCORE}$.
    \item \textbf{Lexical diversity.} Let $\mathcal{V}$ be the set of unique assistant tokens and $\mathcal{T}$ be the multiset of all assistant tokens. We compute
    \[
        r_{\text{lex}} = \frac{|\mathcal{V}|}{|\mathcal{T}|}.
    \]
    We discard $d$ if $r_{\text{lex}} < \texttt{MIN\_LEX\_DIV}$.
\end{itemize}
In addition, dialogues with fewer than $\texttt{MIN\_ASST\_TOTAL\_TOKS}$ assistant tokens in total are removed to avoid overly short conversations.

\paragraph{Scoring and selection.}
For each dialogue that passes all constraints, we compute a normalized heuristic quality score:
\[
    s(d) = 0.45\,(1-r_{\text{short}}) + 0.35\,(1-r_{\text{rep}}) + 0.20\,r_{\text{lex}},
\]
and clip $s(d)$ to $[0,1]$. Finally, we rank all retained dialogues by $s(d)$ (descending) and select the \mbox{top-$10K$} dialogues. This procedure yields a simple, fully deterministic baseline that prioritizes non-trivial, less repetitive, and lexically diverse assistant behavior while requiring no learned model.

\end{document}